\documentclass[10pt,twocolumn,letterpaper]{article}

\usepackage{cvpr}
\usepackage{times}
\usepackage{epsfig}
\usepackage{graphicx}
\usepackage{amsmath}
\usepackage{amssymb}

\usepackage[utf8]{inputenc} 
\usepackage[T1]{fontenc}    
\usepackage{url}            
\usepackage{booktabs}       
\usepackage{amsfonts}       
\usepackage{nicefrac}       
\usepackage{microtype}      

\usepackage{caption}
\usepackage{multirow}
\usepackage{xspace}
\usepackage{mathtools}
\usepackage{wasysym}
\usepackage{marvosym}
\usepackage{xcolor}
\usepackage{caption}
\usepackage{hyperref}
\hypersetup{breaklinks=true,colorlinks,pagebackref=true}

\cvprfinalcopy 

\def\cvprPaperID{3308} 

\def\bx{{\bf x}}
\def\bz{{\bf z}}
\def\ba{{\bf a}}
\makeatother

\makeatletter
\def\blfootnote{\xdef\@thefnmark{}\@footnotetext}
\makeatother

\def\tfidf{TF$\cdot$IDF}

\def\nclips{7637 }
\def\nmovies{51 }

\title{\emph{MovieGraphs}: Towards Understanding Human-Centric Situations from Videos}

\author{Paul Vicol$^{1,2}$ \quad Makarand Tapaswi$^{1,2}$ \quad Llu\'is Castrej\'on$^{3}$ \quad Sanja Fidler$^{1,2}$\\
$^1$University of Toronto \quad $^2$Vector Institute \quad $^3$Montreal Institute for Learning Algorithms\\
{\tt\small \{pvicol, makarand, fidler\}@cs.toronto.edu, lluis.enric.castrejon.subira@umontreal.ca}
\\
{\normalsize \url{http://moviegraphs.cs.toronto.edu}}
}

\begin{document}

\twocolumn[{
\renewcommand\twocolumn[1][]{#1}
\maketitle
\vspace*{-0.5cm}
\centering

  \includegraphics[width=0.9\linewidth]{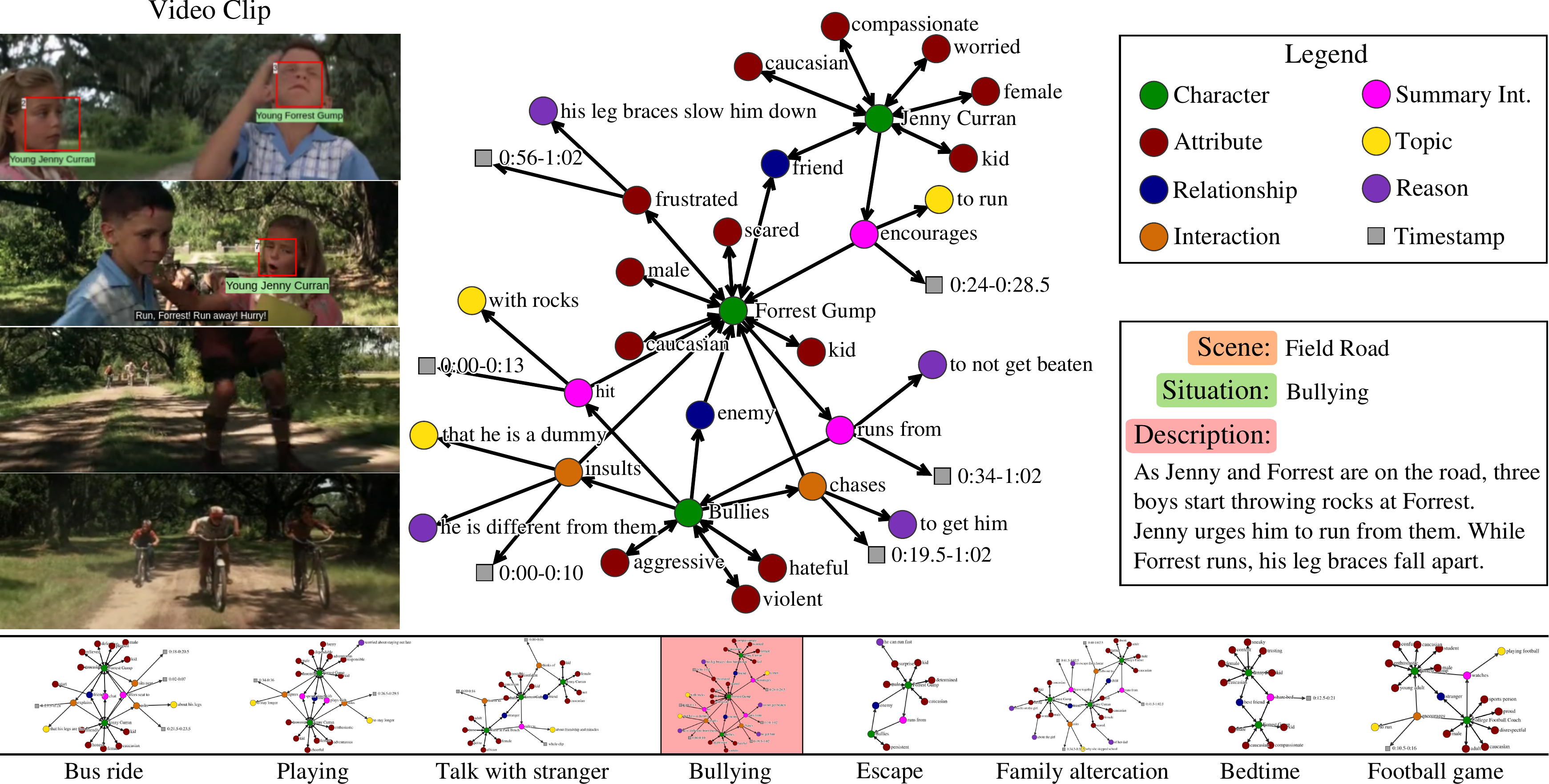}

\vspace*{-0.2cm}
\captionof{figure}{\small An example from the \emph{MovieGraphs} dataset. Each of the \nclips video clips is annotated with:
1) a \textit{graph} that captures the characters in the scene and their attributes, interactions (with topics and reasons), relationships, and time stamps;
2) a \textit{situation label} that captures the overarching theme of the interactions;
3) a \textit{scene label} showing where the action takes place; and
4) a natural language \textit{description} of the clip.
The graphs at the bottom show situations that occur before and after the one depicted in the main panel.}
\label{fig:openingexample}
\vspace*{0.5cm}
}]

\vspace{-2mm}
\begin{abstract}
\vspace{-2mm}
There is growing interest in artificial intelligence to build socially intelligent robots.
This requires machines to have the ability to ``read'' people's emotions, motivations, and other factors that affect behavior. Towards this goal, we introduce a novel dataset called MovieGraphs which provides detailed, graph-based annotations of social situations depicted in movie clips. Each graph consists of several types of nodes, to capture who is present in the clip, their emotional and physical attributes, their relationships (i.e., parent/child), and the interactions between them. Most interactions are associated with \emph{topics} that provide additional details, and \emph{reasons} that give motivations for actions. In addition, most interactions and many attributes are \emph{grounded} in the video with time stamps.
We provide a thorough analysis of our dataset, showing interesting common-sense correlations between different social aspects of scenes, as well as across scenes over time. We propose a method for querying videos and text with graphs, and show that: 1) our graphs contain rich and sufficient information to summarize and localize each scene; and 2) subgraphs allow us to describe situations at an abstract level and retrieve multiple semantically relevant situations. We also propose methods for interaction understanding via ordering, and reason understanding. MovieGraphs is the first benchmark to focus on \textit{inferred} properties of human-centric situations, and opens up an exciting avenue towards socially-intelligent AI agents.
\end{abstract}

\vspace{-4mm}
\section{Introduction}
\label{sec:intro}

An important part of effective interaction is behaving appropriately in a given situation.
People typically know how to talk to their boss, react to a worried parent or a naughty child, or cheer up a friend.
This requires proper reading of people's emotions, understanding their mood, motivations, and other factors that affect behavior.
Furthermore, it requires understanding social and cultural norms, and being aware of the implications of one's actions.
The increasing interest in social chat bots and personal assistants~\cite{herb,tay,Westlund16,Li16,Mataric17,Vinyals15} points to the importance of teaching artificial agents to understand the subtleties of human social interactions.

Towards this goal, we collect a novel dataset called \textit{MovieGraphs} (Fig.~\ref{fig:openingexample}) containing movie clips that depict human-centric situations.  Movies are a rich source of information about behavior, because like people in the real world, movie characters face a variety of situations: they deal with colleagues at work, with family at home, with friends, and with enemies.
Past situations lead to new situations, relationships change over time, and we get to see the same character experience emotional ups and downs just as real people do.
The behavior of characters depends on their interpersonal relationships (\eg~family or friends), as well as on the social context, which includes the scene (\eg~bar) and situation (\eg~date). We use \textit{graphs} to describe this behavior because graphs are more structured than natural language, and allow us to easily ground information in videos.

The \emph{MovieGraphs} dataset consists of \nclips movie clips annotated with graphs
that represent who is in each clip, the interactions between characters, their relationships, and various visible and inferred properties such as the reasons behind certain interactions.
Each clip is also annotated with a situation label, a scene label (where the situation takes place), and a natural language description.
Furthermore, our graphs are visually and temporally grounded: characters in the graph are associated with face tracks in the clip, and most interactions are associated with the time intervals in which they occur.

We provide a detailed analysis of our dataset, showing interesting common-sense correlations between different social aspects of situations.
We propose methods for graph-based video retrieval, interaction understanding via ordering, and understanding motivations via reason prediction.
We show that graphs contain sufficient information to localize a video clip in a dataset of movies, and that querying via subgraphs allows us to retrieve semantically meaningful clips.
Our dataset and code will be released ({\small \url{http://moviegraphs.cs.toronto.edu}}), to inspire future work in this exciting domain.

The rest of this paper is structured as follows:
in Sec.~\ref{sec:related}, we discuss related work;
Sec.~\ref{sec:dataset} describes our dataset;
Sec.~\ref{sec:method} introduces the models we use for video retrieval, interaction ordering, and reason prediction;
Sec.~\ref{sec:results} presents the results of our experiments;
and we conclude in Sec.~\ref{sec:conc}.

\section{Related Work}
\label{sec:related}

\paragraph{Video Understanding.} There is increasing effort in developing video understanding techniques that go beyond classifying actions in short video snippets~\cite{Laptev:2008hb,marszalek09}, towards parsing more complex videos~\cite{Alayrac16,saxena15,Sigurdsson16}.
A large body of work focuses on identifying characters in movies or TV series~\cite{Bojanowski:2013bg,Everingham2006,Ramanathan:2014fj,Tapaswi2012_PersonID} and estimating their poses~\cite{Eichner12}.
Steps towards understanding social aspects of scenes have included classifying four visual types of interactions 
\cite{PatronPerez12}, and predicting whether people are looking at each other~\cite{Jimenez14}.
\cite{Ding10,Park11} find communities of characters in movies and analyze their social networks.
In~\cite{Fathi12}, the authors predict coarse social interaction groups (\eg~monologue or dialog) in ego-centric videos collected at theme parks.
In the domain of affective computing, the literature covers user studies of social behavior~\cite{Jaques16}.
However, we are not aware of any prior work that analyzes and models human-centric situations at the level of detail and temporal scale that we present here.
Additionally, our annotations are richer than in Hollywood2~\cite{hollywood2site} (action labels vs interaction graphs), and more detailed than Large Scale Movie Description Challenge (LSMDC)~\cite{LSDMC} (single sentence vs short descriptions).

\vspace{-3mm}
\paragraph{Video Q\&A.} Other ways to demonstrate video understanding include describing short movie clips~\cite{Rohrbach15,Tapaswi2015_Book2Movie,torabi2015mvad,ZhuICCV15} and answering questions about them~\cite{TGIFQA,pororoQA,MarioQA,Tapaswi2016_MovieQA}.
However, these models typically form internal representations of actions, interactions, and emotions, and this implicit knowledge is not easy to query. We believe that graphs may lead to more interpretable representations.

\vspace{-3mm}
\paragraph{Graphs as Semantic Representations.}
Recently, there has been increasing interest in using graphs as structured representations of semantics.
Johnson~\etal~\cite{johnson2015image} introduce \textit{scene graphs} to encode the relationships between objects in a scene and their attributes, and show that such graphs improve image retrieval compared to unstructured text.
Recent work aims to generate such scene graphs from images~\cite{Xu17}.

While retrieval methods using structured prediction exist, ours is the first to use video.
Thus the potentials in our model are very different, as we deal with a different problem: analyzing people.
Our graphs capture human behavior (\eg~\textit{encourages}) that is inferred from facial expressions, actions, and dialog. In contrast, \cite{johnson2015image} deals with spatial relationships between objects in images (\eg~\textit{in front of}).

\vspace{-3mm}
\paragraph{Semantic Role Labeling.}
\cite{Ruiyu17,yatskar2017commonly,yatskar2016imsitu} deal with recognizing situations in images. This task involves predicting the dominant action (verb) as well as the semantic frame, \ie~a set of action-specific roles. However, these works focus on static images with single actions, while we focus on movie clips (videos and dialogs) and tackle different tasks.

\section{The MovieGraphs Dataset}
\label{sec:dataset}

\begin{table}[t]
 \vspace{-1mm}
\centering
\small
\tabcolsep1.2mm
\begin{footnotesize}
\addtolength{\tabcolsep}{4.5pt}
\begin{tabular}{lrrrr}
\toprule
               &   TRAIN &     VAL &    TEST &   TOTAL \\
\midrule
 \# Movies      &   34    &    7    &   10    &   51    \\
 \# Video Clips & 5050    & 1060    & 1527    & 7637    \\
 Desc \#Words   &   35.53 &   34.47 &   34.14 &   35.11 \\
 Desc \#Sents   &    2.73 &    2.45 &    2.91 &    2.73 \\
 Characters    &    3.01 &    2.97 &    2.9  &    2.98 \\
 Interactions  &    3.18 &    2.48 &    3.11 &    3.07 \\
 Summary Int.  &    2    &    2.06 &    2.05 &    2.02 \\
 Relationships &    3.12 &    2.77 &    3.52 &    3.15 \\
 Attributes    &   13.59 &   14.53 &   13.79 &   13.76 \\
 Topics        &    2.64 &    2.7  &    2.68 &    2.65 \\
 Reasons       &    1.66 &    1.53 &    2.17 &    1.74 \\
 Timestamps    &    4.23 &    4.34 &    4.68 &    4.34 \\
 Avg. Duration &   43.96 &   43.90 &   45.61 &   44.28 \\
\bottomrule
\end{tabular}
\end{footnotesize}

\vspace{-2mm}
\caption{\small
Statistics of the MovieGraphs dataset across train, validation, and test splits.
We show the number of movies and clips; their average duration (sec); the number of words/sentences in descriptions; and average counts of each type of node per graph.
}
\label{table:statstable}
\vspace{-2mm}
\end{table}

We construct a dataset to facilitate machine understanding of real-world social situations and human behaviors. We annotated \nmovies movies; each movie is first split into scenes automatically~\cite{Tapaswi2014_StoryGraphs} and then the boundaries are refined manually such that each clip corresponds to one \textit{social situation}.

We developed a web-based annotation tool that allows human annotators to create graphs of arbitrary size by explicitly creating nodes and connecting them via a drag-and-drop interface. Two key points of our dataset are that each annotator: 1) creates an entire graph per clip, ensuring that each graph is \textit{globally coherent} (i.e., the emotions, interactions, topics make sense when viewed together); and 2) annotates a complete movie, so that the graphs for \textit{consecutive clips} in a movie are also coherent---this would not be possible if annotators simply annotated randomly-assigned clips from a movie.
We provide details on annotation and dataset below.

\subsection{Annotation Interface}
Our annotation interface allows an annotator to view movie clips sequentially. For each clip, the annotator was asked to specify the scene and situation, write a natural language summary, and create a detailed graph of the situation, as depicted in Fig.~\ref{fig:openingexample}. We describe each component of the annotation:

The \textbf{scene label} provides information about the location where the situation takes place, \eg \textit{office, theater, airport}.

The \textbf{situation label} corresponds to the high-level topic of the clip, and summarizes the social interactions that occur between characters, \eg \textit{robbery}, \textit{wedding}.

\begin{figure}[t]
 \vspace{-1mm}
\centering
\includegraphics[width=\linewidth]{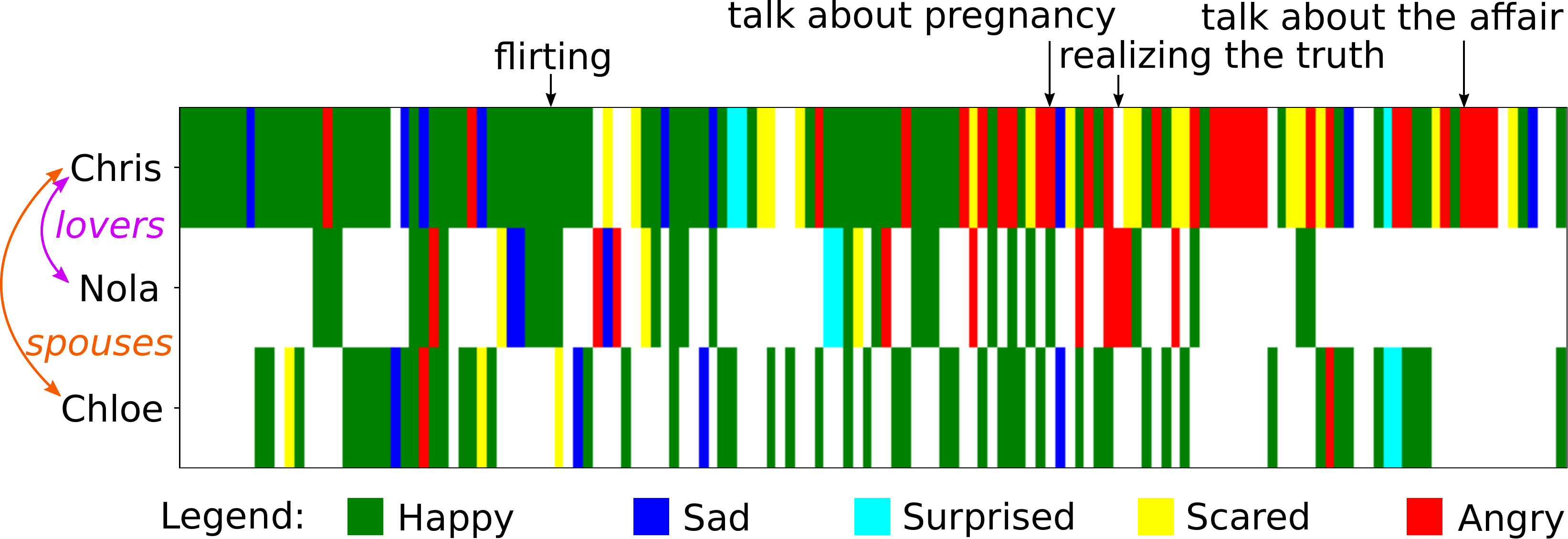}
\vspace{-5mm}
\caption{\small Emotional timelines of the three main characters in ``Match Point.'' The emotions are correlated with situations and relationships between characters.
}
\label{fig:matchpointemotions}
\vspace{-2mm}
\end{figure}

The \textbf{description} provides a multi-sentence, natural language summary of what happens in the clip, based on video, dialog, and any additional information the annotator inferred about the situation.

\begin{figure*}[t]
\vspace{-2mm}
\centering
    \includegraphics[height=5.7cm]{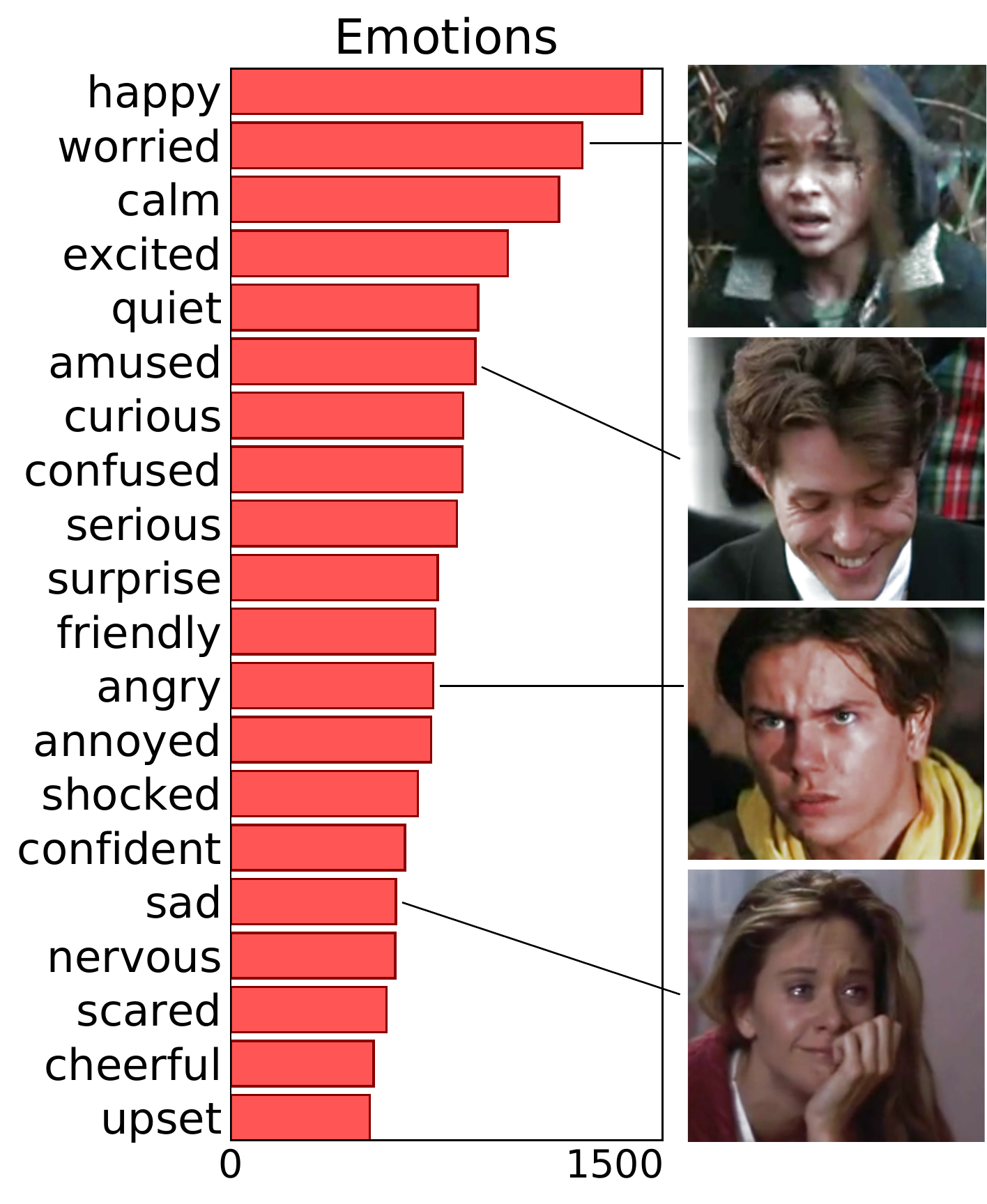}\hspace{6mm}
    \includegraphics[height=5.7cm]{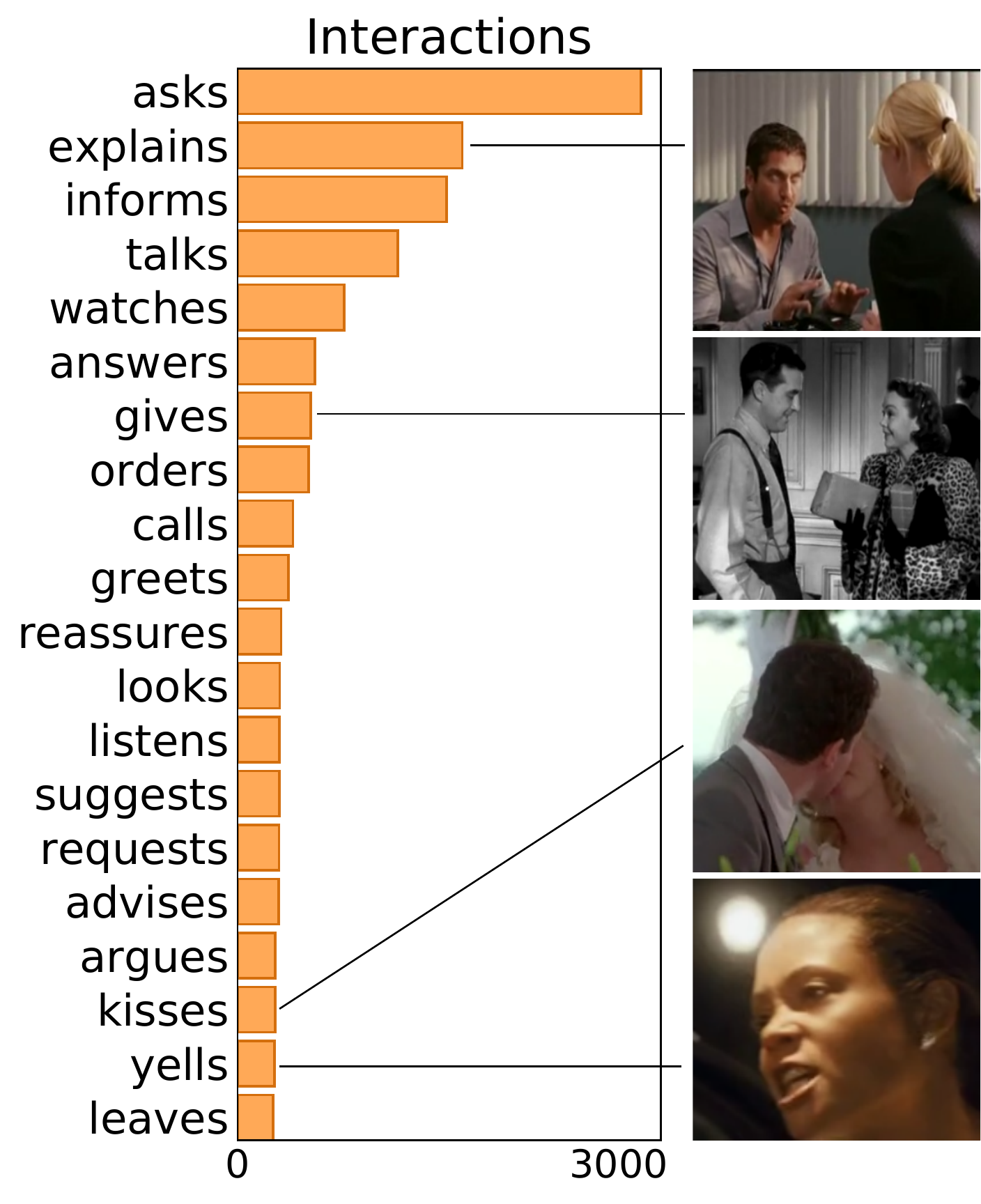}\hspace{6mm}
    \includegraphics[height=5.7cm]{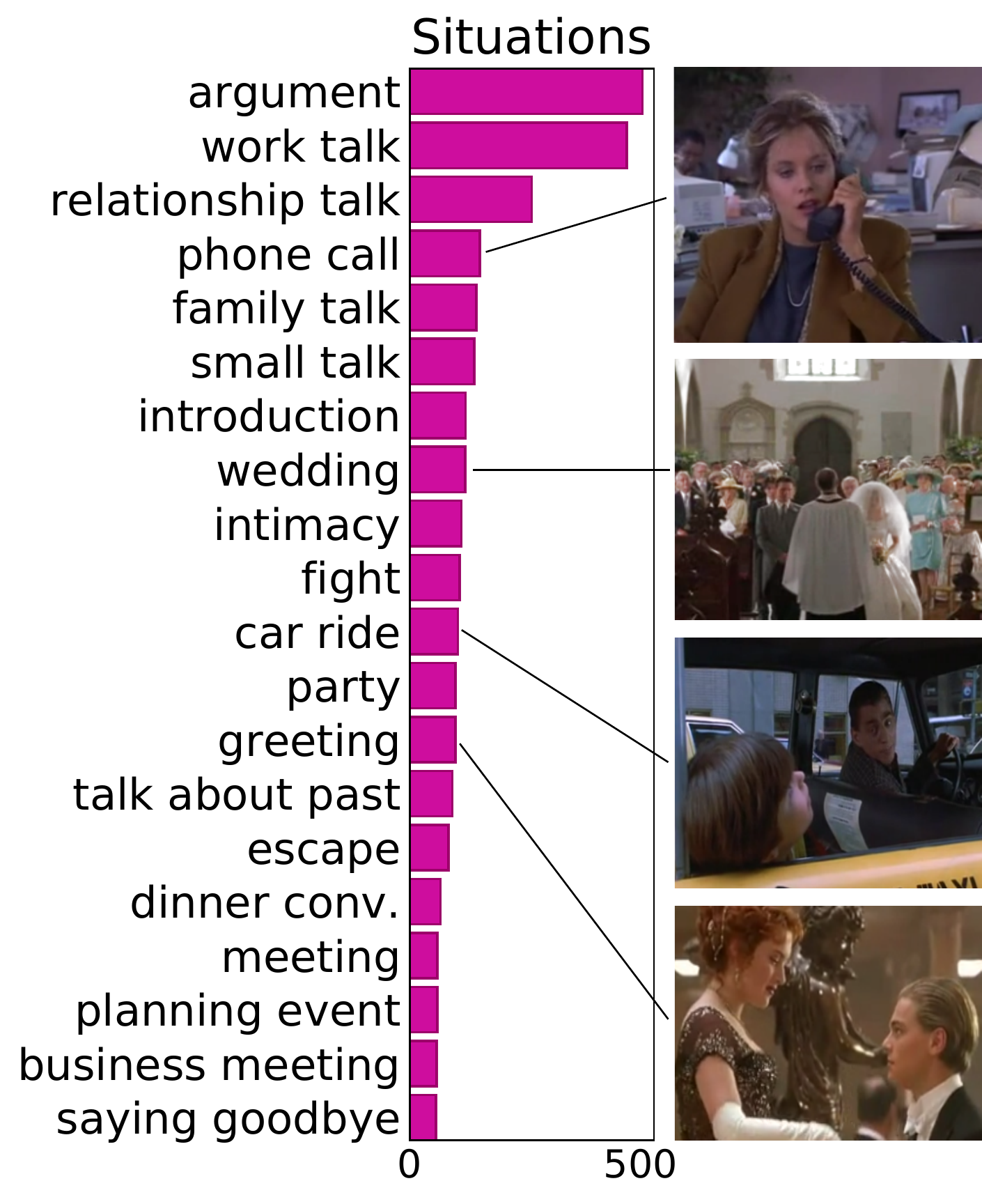}
    \vspace{-2.5mm}
\caption{\small Distributions of the top 20 emotion attributes, interactions, and situations.}
\label{fig:distributions}
\vspace{-2mm}
\end{figure*}

The \textbf{graph} represents a human's understanding of a given situation. Our graphs feature 8 different types of nodes, with edges between them to indicate dependencies. We allow the annotator to choose the \textit{directionality} of each edge. A graph consists of the following node types:

\textbf{Character} nodes represent the people in a scene.
We provide a comprehensive list of character names obtained from IMDb\footnote{\url{http://www.imdb.com/}}, which the annotators can drag and drop onto the graph canvas.

\textbf{Attributes} can be added to character nodes. The categories of attributes are: age, gender, ethnicity, profession, appearance, mental and emotional states.

\textbf{Relationship} nodes can link two or more characters.
The relationships can refer to: family (\eg~\emph{parent}, \emph{spouse}), friendship/romance (\eg~\emph{friend}, \emph{lover}), or work (\eg~\emph{boss}, \emph{co-worker}).
A relationship node can be tagged with a start/end token if it starts or ends in a given clip (\eg the \emph{spouse} relationship starts in a \textit{wedding} clip). Otherwise, we assume that the characters were already in the relationship prior to the scene (\eg already married).

\textbf{Interaction} nodes can be added to link two or more characters.
Interactions can be either verbal (\eg \emph{suggests}, \emph{warns}) or non-verbal (\eg \emph{hugs}, \emph{sits near}).  They can be directed (from one character to another, \eg \emph{A helps B}), or bidirectional if the interaction is symmetric (\eg \emph{A and B argue)}.
A \emph{summary interaction} captures the gist of several local interactions.
Typically there is a single directed summary interaction from each character to the other (\eg \textit{argues}), while there may be many local ones (\eg \textit{asks, replies}).

\textbf{Topic} nodes can be added to interactions to add further details.
For example, the interaction \textit{suggests} may have the topic \textit{to quit the job}.

\textbf{Reason} nodes can be added to interactions and attributes to provide motivations. For example, \textit{apologizes} (interaction) can be linked to \textit{he was late} (reason). Reasons can also be added to emotions: for example, \textit{happy} (emotion) can be linked to \textit{she got engaged} (reason).
Reason nodes contain \textit{inferred} common-sense information. See Table 2 for examples of topics and reasons.

\textbf{Time stamp} nodes ground the graph in the video clip, by providing the time interval in which an interaction or emotional state takes place (\eg a character is \textit{sad}, then (s)he becomes \textit{happy}).

We also perform automatic face tracking, and ask annotators to assign a character name to each track (or mark as false positive). Thus, character nodes are grounded in videos.

\subsection{Data Collection Procedure}
\label{sec:collection}
We hired workers via the freelance website Upwork. We worked closely with a small group of annotators, to ensure high-quality annotations. The workers went through a training phase in which they annotated the same set of clips according to an instruction manual. 
After gathering annotations, we also had a \textit{cross-checking} phase, where annotators swapped movies and checked each others' work.

\subsection{Dataset Statistics}
\label{subsec:statistics}

\begin{figure}[t]
 \vspace{-2mm}
\centering
   \includegraphics[width=0.94\linewidth]{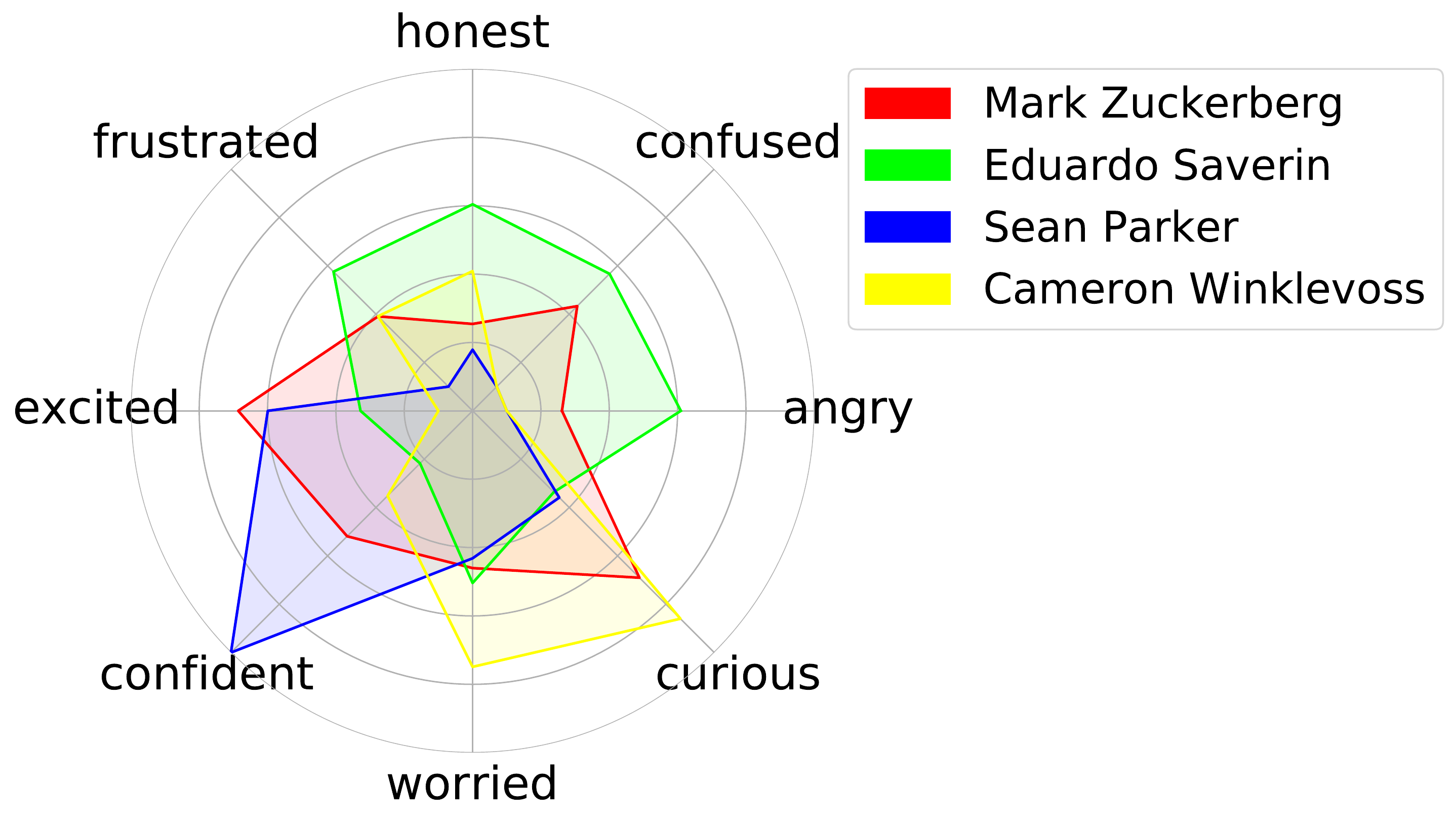}
   \vspace{-2mm}
\caption{\small Emotional profiles from ``The Social Network.''}
\label{fig:socialnetwork_emotions}
 \vspace{-4mm}
\end{figure}

\begin{figure*}[t]
\centering
\raisebox{-0.5\height}{\includegraphics[height=4cm]{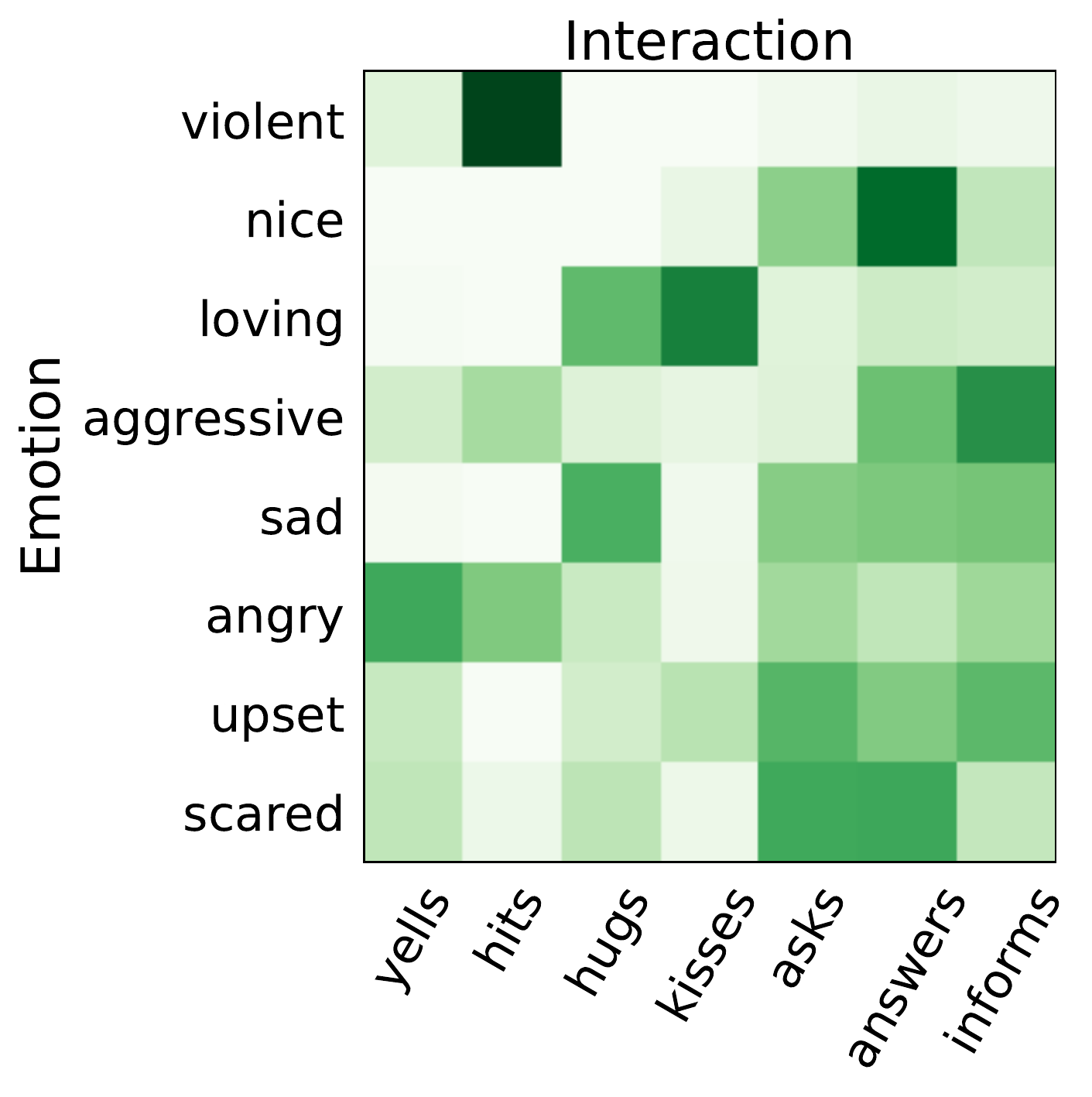}}
\raisebox{-0.5\height}{\includegraphics[height=4cm]{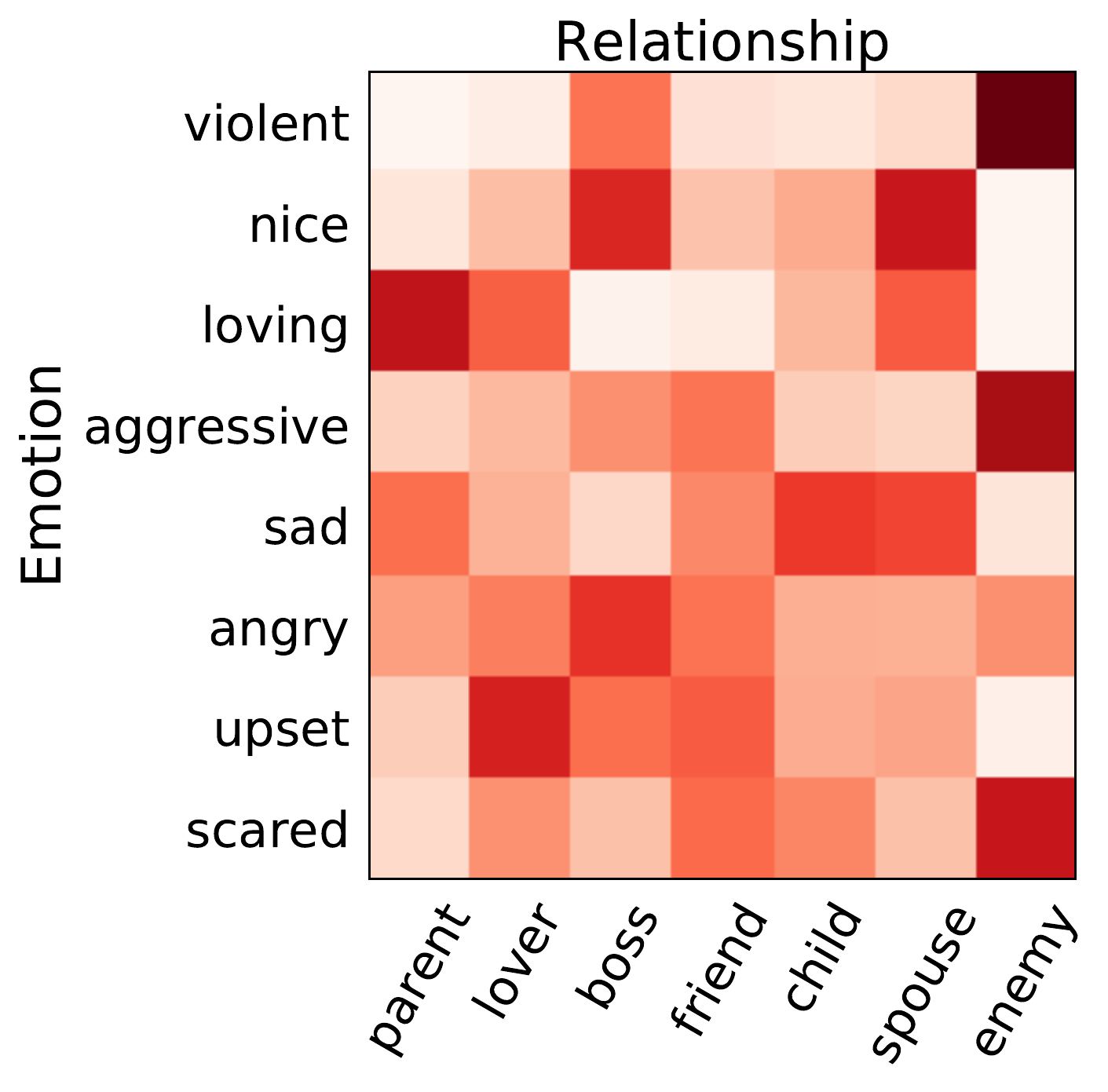}}
\raisebox{-0.5\height}{\includegraphics[height=4cm]{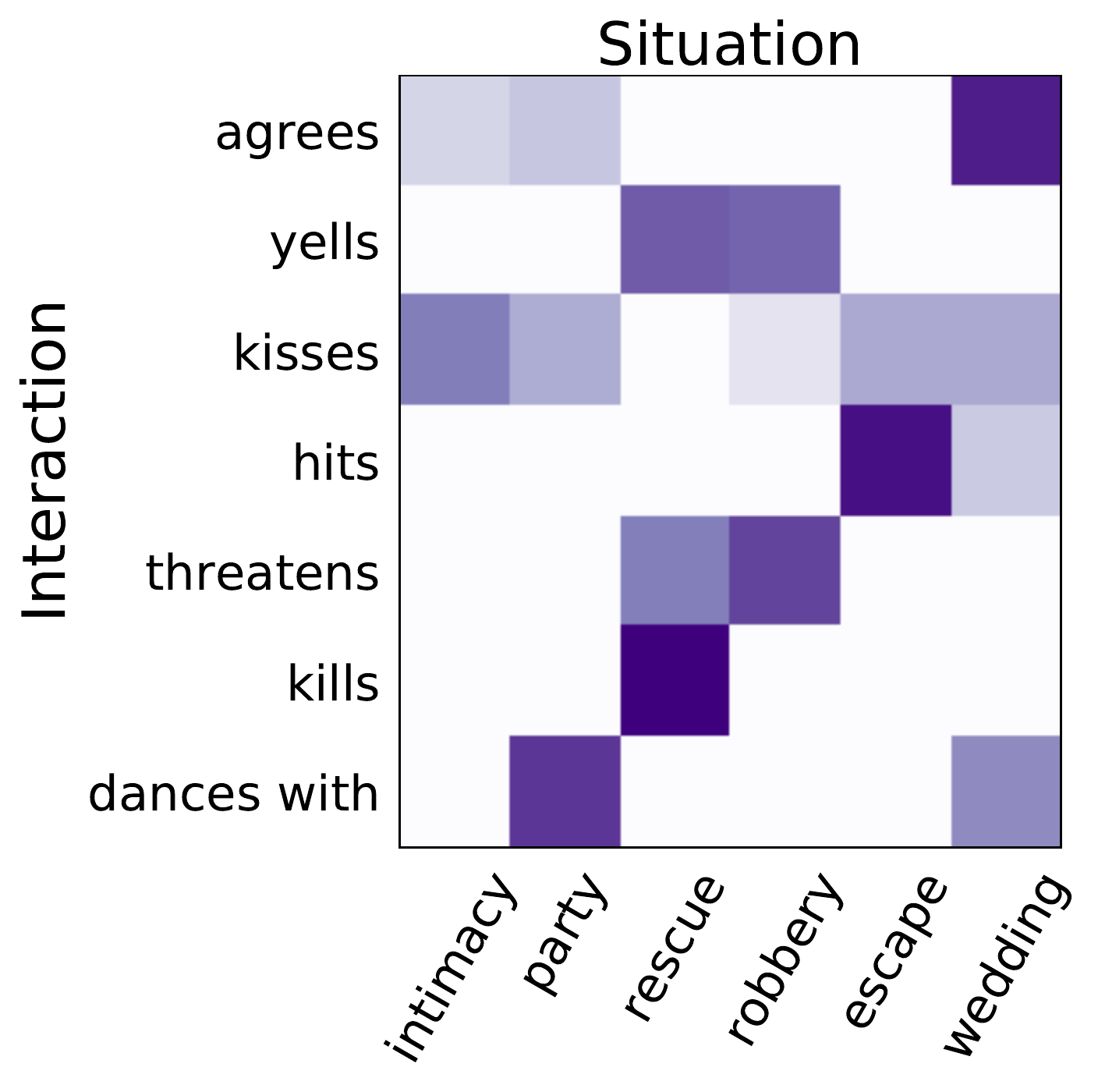}}
\hspace{3mm}
\raisebox{-0.5\height}{\includegraphics[height=3cm]{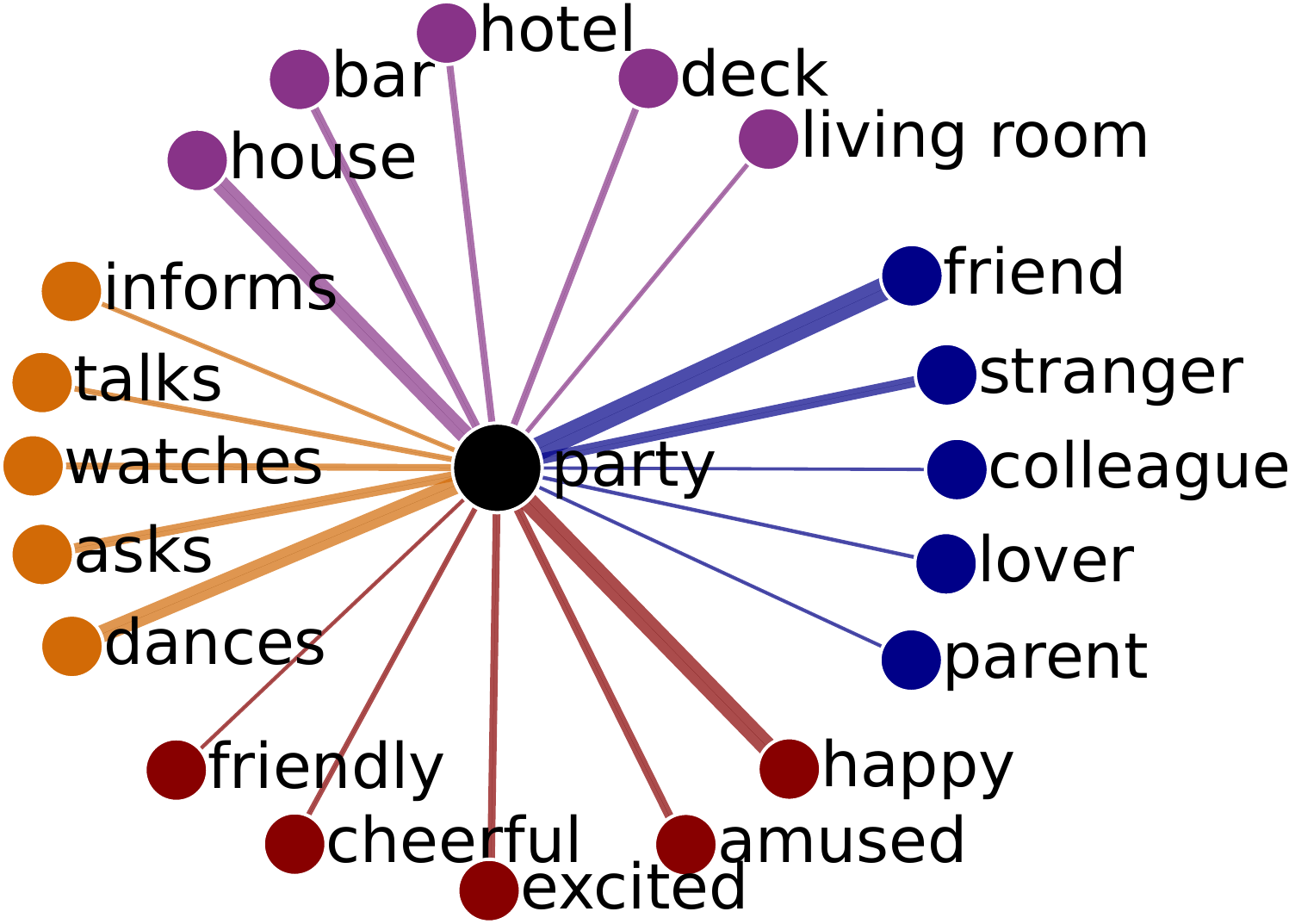}}
 \vspace{-3mm}
\caption{\small {\bf Left:} Correlations between social aspects of scenes. {\bf Right, clockwise from top:} The top-5 most common scenes, relationships, attributes, and interactions associated with the situation \textit{party}.}
\label{fig:Correlations}
\vspace{-2mm}
\end{figure*}

Our dataset consists of \nclips annotated clips from \nmovies movies. Dataset statistics are shown in Table~\ref{table:statstable}.
The majority of the clips contain between 2 and 4 characters, and a graph has on average 13.8 attributes and 3.1 interactions. Fig.~\ref{fig:distributions} shows the distributions of the top 20 emotion attributes, interactions, and situations. We show correlations between node types for a selected set of labels in Fig.~\ref{fig:Correlations}, and the most common social aspects of scenes associated with the situation \textit{party}, to showcase the insight offered by our dataset.

The dataset annotations allow us to follow a character throughout a movie. Fig.~\ref{fig:matchpointemotions} shows the emotions experienced by the three main characters of the movie ``Match Point,'' clip by clip. The emotions make sense when viewed in the context of the situations: when the characters flirt, they are happy; when they talk about problematic issues (pregnancy, the truth, the affair), they are angry. Fig.~\ref{fig:socialnetwork_emotions} shows the emotional profiles of characters from the movie ``The Social Network,'' obtained by aggregating the characters' emotions over all clips.

\begin{figure}
\vspace{-3mm}
\centering
\includegraphics[trim=0 0 0 0,clip=true,angle=-90,origin=c,width=\linewidth,height=5.4cm]{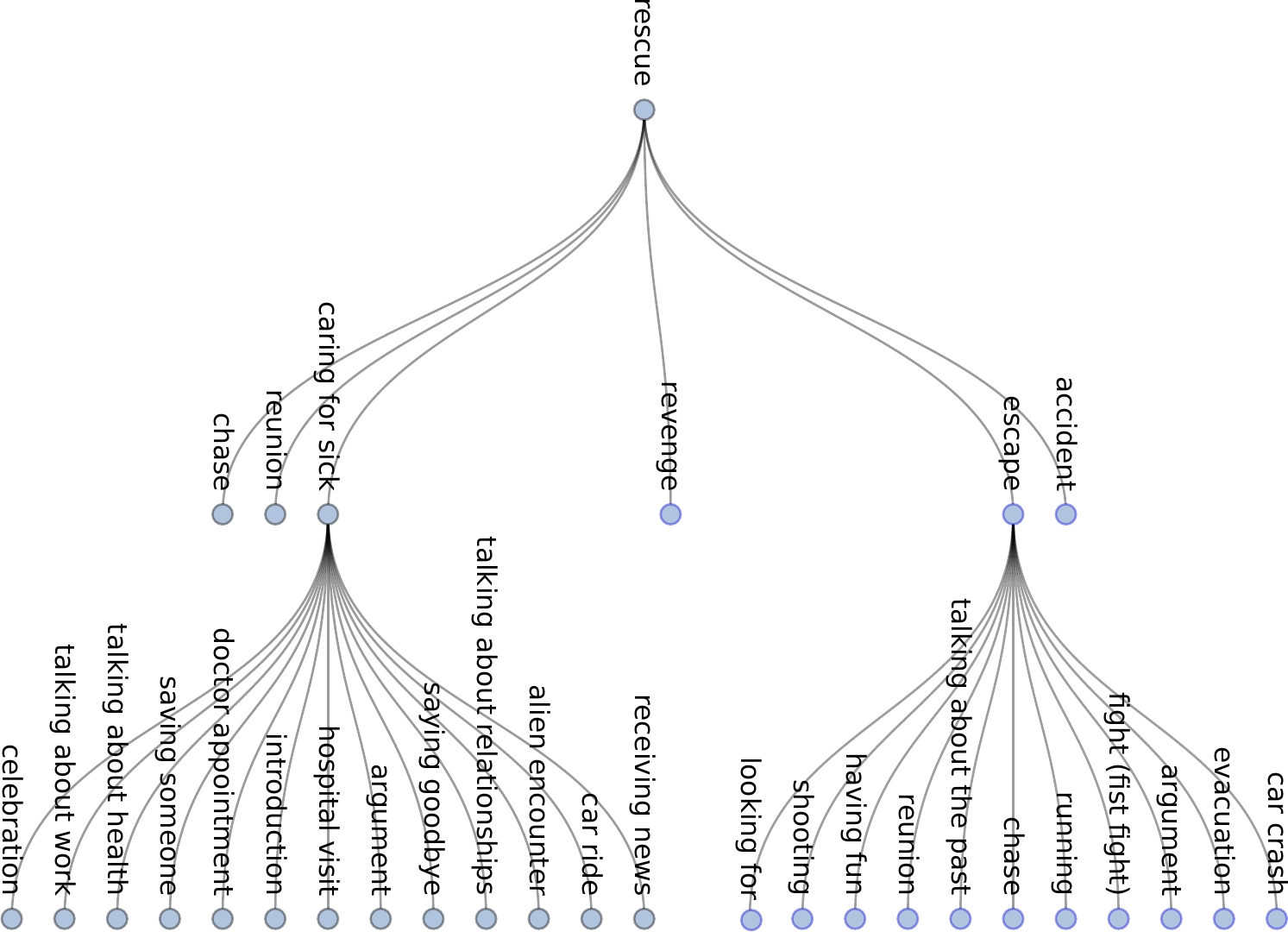}
\vspace{-15mm}
\caption{\small Flow of situations aggregated from all movies.}
\label{fig:situationtree}
\vspace{-4mm}
\end{figure}

In movies, like in real life, situations follow from other situations.
In Fig.~\ref{fig:situationtree}, we present a tree of situations rooted at \textit{rescue}; this is essentially a knowledge graph that shows possible pairwise transitions between situations.


\vspace{-1mm}
\section{Situation Understanding Tasks}
\label{sec:method}

Graphs are an effective tool for capturing the gist of a situation, and are a structured alternative to free-form representations such as textual descriptions.
We propose three tasks to demonstrate different aspects of situation understanding: 1) video clip retrieval using graphs as queries; 2) interaction sorting; and 3) reason prediction.
In this section, we describe these tasks and propose models to tackle them.

\begin{table}[t!]
\vspace{-1mm}
\centering
\label{table:reasonsandtopics}
{\small
\begin{tabular}{@{}ll@{}}
\toprule
\multicolumn{2}{c}{\textbf{Interaction: ``asks''}}               \\ \midrule
\multicolumn{1}{c}{\textbf{Topic}} & \multicolumn{1}{c}{\textbf{Reason}} \\ \midrule
who she is                      & she is pretty                  \\
is this love at first sight     & can't stop looking at her      \\
if he is sober                  & he is the driver               \\
about the speech                & he's the best man              \\
about wedding gifts list        & he needs to buy one            \\
for help                        & he is late again               \\
if he is for bride or groom     & to determine seating area      \\
for time alone                  & to think                       \\
\bottomrule
\end{tabular}
\vspace{-2mm}
\caption{Topics and reasons associated with the interaction ``asks'' in the movie ``Four Weddings and a Funeral.''}
}
\vspace{-2mm}
\end{table}

\subsection{Graph-Based Situation Retrieval}

Here we aim to use graphs as queries to retrieve relevant clips from our dataset, where each clip consists of video and dialog.
We assume our query is a graph $G = (\mathcal{V}, \mathcal{E})$ consisting of different types of nodes $v^{type}\in \mathcal{V}$ and edges between them.
We use the notation $v^{\mathit{ch}}$, $v^{\mathit{att}}$, $v^{\mathit{rel}}$, $v^{\mathit{int}}$, $v^{\mathit{topic}}$, and $v^{\mathit{reason}}$ to denote character, attribute, relationship, interaction, topic, and reason nodes.
Character nodes are the backbone of the graph: all other nodes (except for topics and reasons) link to at least one character node.
To ease notation, we consider the scene and situation labels as disconnected nodes in the graph, $v^{sc}$ and $v^{si}$, respectively.

To perform retrieval, we aim to learn a real-valued function $F_{\theta}(M,G)$ that scores the similarity between a movie clip $M$ and the query graph $G$, where $F_\theta$ should score the highest for the most relevant clip(s).
At test time, we are interested in retrieving the clip with the highest similarity with $G$.
We design $F_{\theta}$ to exploit the structure of the graph and evaluate it against a clip in a semantically meaningful way.
In particular, we reason about the \textit{alignment} between character nodes in the graph and face/person tracks in the video. Given an alignment, we score attributes, interactions, and other nodes accordingly.

Each clip $M$ typically contains several video shots.
We automatically parse each clip to obtain face tracks in each shot, and cluster tracks with similar faces across shots.
To model interactions and relationships, we extend the face detection boxes to create full-body person tracks.
We represent each cluster $c_j$  with a feature vector $\bx_j$, and a pair of clusters $(c_j,c_k)$ with a feature vector $\bx_{jk}$.
Global information about the clip is captured with a feature vector $\bx_{scene}$.
Additional details are provided in Sec.~\ref{subsec:implementation_details}.

We define a random variable $\bz=(z_1,\dots,z_N)$ which reasons about the alignment between character nodes $v_i^{ch}$ and face clusters, $z_i\in\{1,\dots,K\}$.
Here, $N$ is the number of character nodes in the query $G$, and $K$ is the number of face clusters in the clip.
We restrict $\bz$ to map different nodes to different clusters, resulting in all permutations $\bz\in P(K,N)$. In practice, $N = 5$ and $K = 7$.

We define a function that scores a graph in the video clip given an alignment $\bz$ as follows:
\begin{multline}
\label{eq:main_graph_matching}
F_{\theta}(M,G,\bz) = \phi_{sc}(v^{sc}) +  \phi_{si}(v^{si}) \\
+ \sum_i \Big( \phi_{ch}(v_i^{ch},z_i) + \phi_{att}(\mathcal{V}_{i}^{att}, z_i) \Big) \\
+ \sum_{i, j} \phi_{int}(\mathcal{V}_{ij}^{int}, z_i, z_j) + \sum_{i, j} \phi_{rel}(\mathcal{V}_{ij}^{rel}, z_i, z_j).
\end{multline}
The set of attributes associated with character $i$ is $\mathcal{V}_{i}^{att} = \{ v_{k}^{att} : (i, k) \in \mathcal{E} \}$ and the set of interactions between a pair of characters $(i, j)$ is $\mathcal{V}_{ij}^{int} = \{ v_k^{int} : (i, k), (k, j) \in \mathcal{E} \}$, where all edges are directed. The set of relationships is defined similarly.
Here, $\phi$ are potential functions which score components of the graph in the clip.
Each $\phi$ also depends on the clip $M$ and learned parameters $\theta$, which we omit for convenience of notation.

We now describe each type of potential in more detail.
To form the query using the graph, we embed each node label using word embeddings.
For nodes that contain phrases, we mean-pool over the words to get a fixed length representation $\ba^{type}$ (where $type$ is $att$, $int$, etc.).
In our case, $\ba^{type}\in \mathbb R^{100}$ (GloVe~\cite{pennington2014glove}).
We learn two linear embeddings for each type, $W_g^{type}$ for query node labels and $W_m^{type}$ for observations, and score them in a joint space.
We share $W_g$ across all node types to prevent overfitting.

\vspace{-3mm}
\paragraph{Video-Based Potentials.} The attribute unary potential computes the cosine similarity between node embeddings and visual features:
\begin{equation}
\label{eq:unaries}
\phi_{att}(\mathcal{V}_i^{att}, z_i) = \Big\langle W_g \sum_{\mathclap{v_k \in \mathcal{V}_i^{att}}} \ba_k^{att}, W_m^{att} \bx_{z_i}^{att} \Big\rangle \, .
\end{equation}
A similar potential is used to score the scene $v^{sc}$ and situation $v^{si}$ labels with video feature $\bx_{scene}$  (but does not depend on $\bz$).
Furthermore, we score pairwise dependencies as:
\begin{equation}
\label{eq:pairwise}
\phi_{type}(\mathcal{V}_{ij}^{type}, z_i,z_j) = \Big\langle W_g \sum_{\mathclap{v_k \in \mathcal{V}_{ij}^{type}}} \ba_k^{type}, W_m^{type} \bx_{z_i z_j} \Big\rangle
\end{equation}
with $\mathit{type} \in \{ \mathit{rel}, \mathit{int} \}$.

\vspace{-4mm}
\paragraph{Scoring Dialog.}
To truly understand a situation, we need to consider not only visual cues, but also dialog.
For this, we learn a function $Q$ to score a query $G$ with dialog $D$ as:
\begin{equation}
\label{eq:dlg}
Q(D, G) = \sum_{v_k \in \mathcal{V}} \sum_{i} \max_j ((W_g \ba_{k,i})^T(W_d \bx_{d_j})) \, ,
\vspace{-2mm}
\end{equation}
where $\ba_{k,i}$ is the GloVe embedding of the $i^{\mathrm{th}}$ word in node $v_k$, and $\bx_{d_j}$ is the
embedding of the $j^{\mathrm{th}}$ dialog word. This finds the best matching word in the dialog for each word in the graph, and computes similarity by summing across all graph words.
We initialize the matrices $W_g$ and $W_d$ to identity, because GloVe vectors already capture relevant semantic information.
To take into account both video and dialog, we perform late fusion of video and dialog scores (see Sec.~\ref{subsec:exp_retrieval}).

\vspace{-4mm}
\paragraph{Person Identification.}
To classify each face cluster as one of the characters, we harvest character and actor images from IMDb.
We fine-tune a VGG-16~\cite{parkhi2015deep} network on these images, combined with our video face crops, using a triplet loss (i.e., minimizing the Euclidean distance between embeddings of two positive examples wrt  a negative pair).
To compute $\phi_{ch}(v_i^{ch}, z_i)$, we find the embedding distance between the face track and each movie character, and convert it into a probability.
For details, see Suppl. Mat.~\ref{supp:sec:personid}.

\vspace{-3mm}
\subsubsection{Learning and Inference}
\label{subsec:learning}
\paragraph{Learning.}
Our training data consists of tuples $(G_n, M_n, \bz_n)$: for each graph we have an associated clip and ground-truth alignment to face clusters.
We learn the parameters of $F_\theta$ using the max-margin ranking loss:
\begin{equation}
\label{eq:comparing_subgraphs_loss}
\mathcal{L}_\theta = \sum_{\mathclap{(n,n')}} \max(0, 1 - (F_\theta(G_n,M_n,\bz_n) - F_\theta(G_n, M_{n'}, \bz_{n'}))) \, ,
\end{equation}
where $n'$ is an index of a negative example for $G_n$.
In practice, we sample three classes of negatives: 1) clips from other movies (different characters, therefore easy negatives); 2) different clips from the same movie (medium difficulty); or 3) the same clip with different alignments $\bz_{n'}$ (same characters, aligned with the clip incorrectly, therefore hard negatives).
We train the dialog model $Q(D, G)$ similarly, with a max-margin ranking loss that does not involve $\bz$.
We use the Adam optimizer~\cite{kingma2014adam} with learning rate 0.0003.


\vspace{-4mm}
\paragraph{Inference.}
We perform an exhaustive search over all clips and alignments to retrieve the most similar clip for the query graph $G$:
\begin{equation}
M^*= \arg\max_n \big(\max_\bz F_{\theta}(G, M_n, \bz) \big) \, .
\end{equation}

\vspace{-1mm}
\subsubsection{Implementation Details}
\label{subsec:implementation_details}
\vspace{-1mm}

\paragraph{Video Features.}
To obtain a holistic video representation, we process every fifth frame of the video using the Hybrid1365-VGG model~\cite{zhou2016places} and extract \texttt{pool5} features.
We mean pool over space and time to obtain one representation $\bx_{scene} \in \mathbb{R}^{512}$ for the entire clip.
For each face cluster, we compute age and gender predictions~\cite{LH:CVPRw15:age} (Eq.~\ref{eq:unaries}, $\bx_{z_i}^{age}$, $\bx_{z_i}^{gen}$) and extract features from another CNN trained to predict emotions~\cite{LH:ICMI15:emo} (Eq.~\ref{eq:unaries}, $\bx_{z_i}^{att}$).
This allows us to score unary terms involving attributes.

We extend the face detections to obtain person detections and tracks that are used to score pairwise terms.
We represent each person track by pooling features of spatio-temporal regions in which the person appears.
Specifically, $\bx_{z_i z_j}$ (Eq.~\ref{eq:pairwise}) is computed by stacking such person track features $[\bx_{z_i}^p; \bx_{z_j}^p]$.
Note that ordered stacking maintains edge directions ($v^{ch}_i \rightarrow v^{int,rel} \rightarrow v^{ch}_j$).

\vspace{-5mm}
\paragraph{Text.}
We evaluate two representations for text modalities:
(i) \tfidf~\cite{Manning2008}, where we use the logarithmic form; and
(ii) GloVe~\cite{pennington2014glove} word embeddings.
Similar to~\cite{Tapaswi2016_MovieQA}, scoring the dialogs with \tfidf~involves representing the graph query and dialog text as sparse vectors ($\mathbb{R}^{|\mathrm{vocab}|}$) and computing their cosine similarity.
We explore two pooling strategies with word embeddings:
1) \emph{max-sum} (Eq.~\ref{eq:dlg}); and
2) \emph{max-sum $\cdot$ idf}, which weighs words based on rarity.

\vspace{-1mm}
\subsection{Interaction Ordering}
\label{subsec:interaction_ordering}

Predicting probable future interactions on the basis of past interactions, their topics, and the social context is a challenging task.
We evaluate interaction understanding via the proxy task of learning to \textit{sort} a set of interactions into a plausible order (Table~\ref{table:interactionsorting}).
We present a toy task wherein we take the interactions between a pair of characters, and train an RNN to \textit{choose interactions} sequentially from the set, in the order in which they would likely occur.
We represent an interaction and corresponding topic by the concatenation of their GloVe embeddings, with an additional digit appended to indicate the direction of the interaction.
We train an attention-based decoder RNN to regress interaction representations: at each time step, it outputs a vector that should be close to the embedding of the interaction at that step in the sequence. We use a single-layer GRU~\cite{cho2014GRU}, and condition on a 100-d context vector formed by applying linear layers on the situation, scene, relationship, and attribute embeddings.
We zero-mask one interaction from the input set at each time step, to ensure that the model does not select the same interaction multiple times. Masking is done with teacher-forcing during training, and with the model's predictions at test time.
For details, see Suppl. Mat.~\ref{supp:sec:interaction_ordering}.

\subsection{Reason Prediction}
\label{subsec:reason_prediction}

Given information about the scene in the form of attributes of each character, their relationship, and an interaction in which they are engaging, we aim to predict plausible reasons for why the interaction took place. Scene and situation labels are also used, to provide global context.

As in previous tasks, we first represent the relevant nodes by their GloVe embeddings.
The characters are identity agnostic and are represented as a weighted combination of their attributes.
We encode individual components of the sub-graph (scene, situation, interaction, relationship) through linear layers and learn a 100-d context vector.

Our decoder is a single-layer GRU with 100 hidden units that conditions on the scene description (context vector), and produces a reason, word by word.
As is standard, the decoder sees the previous word and context vector at each time step to generate the next word.
To obtain some variability during sampling, we set the temperature to 0.6.
We train the model end-to-end on the train and val sets (leaving out a few samples to choose a checkpoint qualitatively), and evaluate on test.
Please refer to Suppl. Mat.~\ref{supp:sec:reason_pred} for details.

\section{Experimental Results}
\label{sec:results}

\begin{figure*}
\vspace{-2mm}
\centering
    \includegraphics[height=3.7cm]{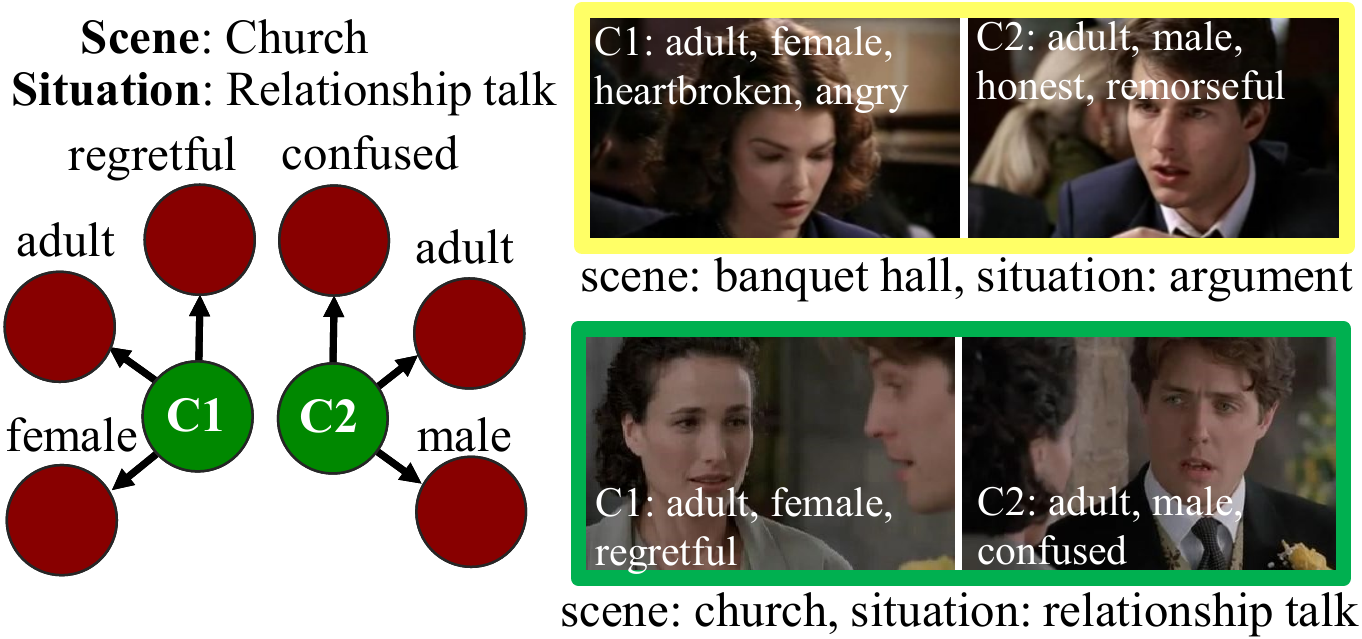} \hspace{0.5cm}
    \includegraphics[height=3.7cm]{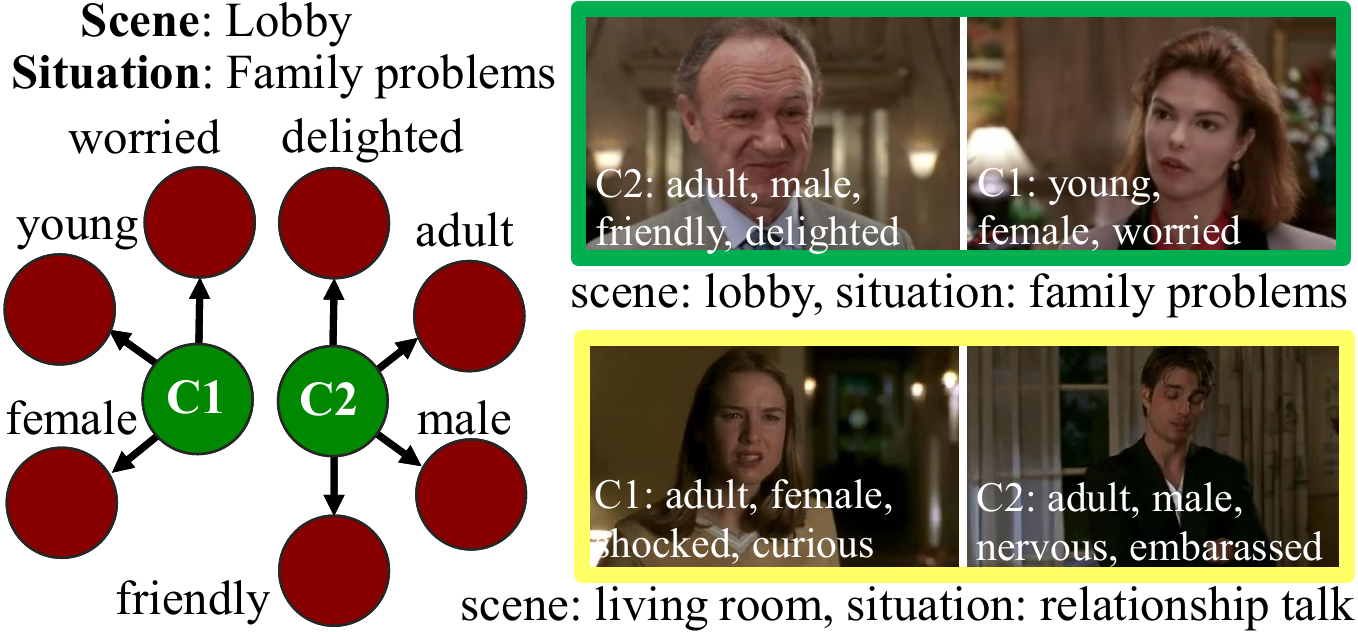}
\vspace{-2mm}
\caption{
Identity agnostic sub-graph queries and the top-2 retrieved clips, which are from different movies.
We search for video clips that have overall similarity with respect to scene and situation, and also character attributes and emotions.
The yellow boxes indicate results that are quite similar in meaning to the query, and the green boxes indicate ground-truth.}
\label{fig:retrieval_examples}
 \vspace{-4mm}
\end{figure*}

The \emph{MovieGraphs} dataset is split into \emph{train}, \emph{val}, and \emph{test} sets with a 10:2:3 ratio of clips (see Table~\ref{table:statstable}), and no overlapping movies.
We learn model parameters on \emph{train}, choose checkpoints on \emph{val}, and present final evaluation on \emph{test}.

\vspace{-4mm}
\paragraph{Face Clustering and Person Identification.}
On average, our clips have 9.2 valid face tracks which form 2.1 ground-truth clusters.
For face clustering, we obtain a weighted clustering purity of 75.8\%,
which is reasonable, as we do not filter background characters or false-positive tracks.
Person identification (ID) for a large number of movies spanning many decades is hard, due to the differences between IMDb gallery images and video face tracks.
We obtain a track-level identification accuracy of 43.7\% vs. chance at 13.2\%.
We present details in Suppl. Mat.~\ref{supp:sec:personid}.

\vspace{-1mm}
\subsection{Graph-based Retrieval}
\label{subsec:exp_retrieval}


\begin{table}[t]
\centering
\tabcolsep=0.15cm
\addtolength{\tabcolsep}{-0.7pt}
{\footnotesize

\begin{tabular}{l l | c c | c c c c}
\toprule

& \multirow{2}{*}{Method} & \multicolumn{2}{c}{PersonID} & \multicolumn{4}{|c}{TEST} \\

& & CL & ID & R@1 & R@5 & R@10 & med.-R \\
\midrule

 1 & random, movie unkn.                         &     -     &     -     & 0.1 &  0.3 &  0.7 & 764 \\
 2 & random, movie known                         &     -     &     -     & 0.7 &  3.3 &  6.6 & 78  \\
\midrule

\multicolumn{7}{c}{\bf DESCRIPTION} \\
 3 & \tfidf                           &  -  &  -  & 61.6 & 83.8 & 89.7 &  1    \\
 4 & GloVe, max-sum                   &  -  &  -  & 62.1 & 81.3 & 87.2 &  1    \\
 5 & GloVe, idf $\cdot$ max-sum       &  -  &  -  & 61.3 & 81.6 & 86.9 &  1    \\
\midrule

\multicolumn{7}{c}{\bf DIALOG} \\
 6 & \tfidf                           &  -  &  -  & 31.8  & 49.8 & 57.2 & 6    \\
 7 & GloVe, max-sum                   &  -  &  -  & 28.0  & 42.4 & 50.2 & 10   \\
 8 & GloVe, idf $\cdot$ max-sum       &  -  &  -  & 28.7  & 43.1 & 50.2 & 10   \\
 \midrule

\multicolumn{7}{c}{\bf MOVIE CLIP} \\
 9 & sc                                          &  -   &   -  &  1.1 &  4.3 &  7.7 & 141.5 \\
10 & sc, si                                      &  -   &   -  &  1.0 &  5.4 &  8.7 & 140   \\
11 & sc, si, att                                 & pr   & pr   &  2.2 &  9.4 & 15.5 & 84    \\
12 & sc, si, att, rel, int                       & pr   & pr   &  2.7 & 10.9 & 18.9 & 59    \\
13 & sc, si, att, rel, int                       & pr   & gt   &  7.7 & 28.8 & 44.9 & 13    \\
14 & sc, si, att, rel, int                       & gt   & gt   & 13.0 & 37.4 & 50.4 & 10    \\
15 & sc, si, att, rel, int, dlg                  & pr   & pr   & 31.6 & 50.4 & 56.6 & 5    \\
16 & sc, si, att, rel, int, dlg                  & gt   & gt   & 40.4 & 62.1 & 71.1 & 3    \\

\bottomrule
\end{tabular}}
\vspace{-2mm}
\caption{\small
Retrieval results when using the graph as a query. \emph{dlg} refers to dialog.
For PersonID, CL and ID indicate clustering and identification; \emph{gt}~denotes ground-truth, and \emph{pr}~denotes predictions.}
\label{table:all_retrieval}
\vspace{-4mm}
\end{table}

All retrieval results are shown in Table~\ref{table:all_retrieval}.
Similar to image-text retrieval (\eg~Flickr8k~\cite{Hodosh2013Flickr8k}), we use the following metrics: median rank and recall at K (1, 5, 10).
Unless mentioned otherwise, we assume that the entire graph is used as part of the query.
The first two rows show the performance of a random retrieval model that may or may not know the source movie.

\vspace{-4mm}
\paragraph{Description Retrieval.}
Our first experiment evaluates the similarity between graphs and clip descriptions.
We use the three models described in Sec.~\ref{subsec:implementation_details}: \tfidf, \emph{max-sum}, and \emph{max-sum $\cdot$ idf}.
We consistently obtain median rank 1 (Table~\ref{table:all_retrieval}, rows 3-5), possibly due to descriptive topics and reasons, and character names that help localize the scene well (see Suppl. Mat.~\ref{supp:sec:tfidfablativestudy} for an ablation study on node types).

\vspace{-5mm}
\paragraph{Dialog Retrieval.}
In our second experiment, we aim to retrieve a relevant clip based on dialog, given a graph.
This is considerably harder, as many elements of the graph are visual (\eg~\textit{kisses}) or inferred from the conversation (\eg~\textit{encourages}).
We evaluate dialog retrieval with the same models used for descriptions.
Here, GloVe models (rows 7, 8) perform worse than \tfidf~(row 6) achieving med.-R 10 vs 6.
We believe that this is because the embeddings for several classes of words are quite similar, and confuse the model.

\vspace{-4mm}
\paragraph{Movie Clip Retrieval.}
Our third experiment evaluates the impact of visual modalities.
Note that if the query consists only of scene or situation labels, there are multiple clips that are potential matches.
Nevertheless, starting from a random median rank of 764, we are able to improve the rank to 141.5 (row 9) with the scene label only, and 140 with scene and situation labels (row 10).
Directly mapping high-level situations to visual cues is challenging.

Similar to the way characters help localize descriptions and dialogs, person identification helps localization in the visual modality.
If our query consists only of characters, in the best case scenario of using ground-truth (gt) clustering and ID, we obtain a median rank of 17.
Our predicted (pr) clustering works quite well, and obtains median rank 19.
Owing to the difficulty of person ID, using pr clustering and ID pushes the median rank to 69.

Rows 9-16 present an ablation of graph components.
We start with the scene, situation, attributes, and characters (rows 9-11) as part of our query graphs.
Including interactions (+topics) and relationships improves the rank from 84 to 59 (row 12).
In a scenario with gt clusters and ID, we see a large improvement in med.-R from 59 to 10 (rows 13, 14).

\vspace{-5mm}
\paragraph{Late Fusion.}
We combine video and dialog cues using late fusion of the scores from the models used in rows 8 and 12/14, and see a large increase in performance in both pr-pr and gt-gt settings (rows 15, 16).
This points to the benefits offered by dialog in our model.

\vspace{-5mm}
\paragraph{Qualitative Results.}
We present an example of sub-graph retrieval in Fig.~\ref{fig:retrieval_examples}.
Even with small queries (sub-graphs) and identity agnostic retrieval, we obtain interesting results.

\begin{table}[t]
\centering
{\small
\begin{tabular}{c|c}
\toprule
Fully Sorted Accuracy &  Longest Common Subsequence \\
40.5\% (27\%)           &    0.74 (0.67)          \\
\bottomrule
\end{tabular}
}
\vspace{-2mm}
\caption{Performance of our interaction sorting approach on the test set.
The number in ($\cdot$) is random chance.}
\vspace{-1mm}
\label{table:interactionmetrics}
\end{table}

\begin{figure}[]
\vspace{-3mm}
\centering
    \includegraphics[width=0.98\linewidth]{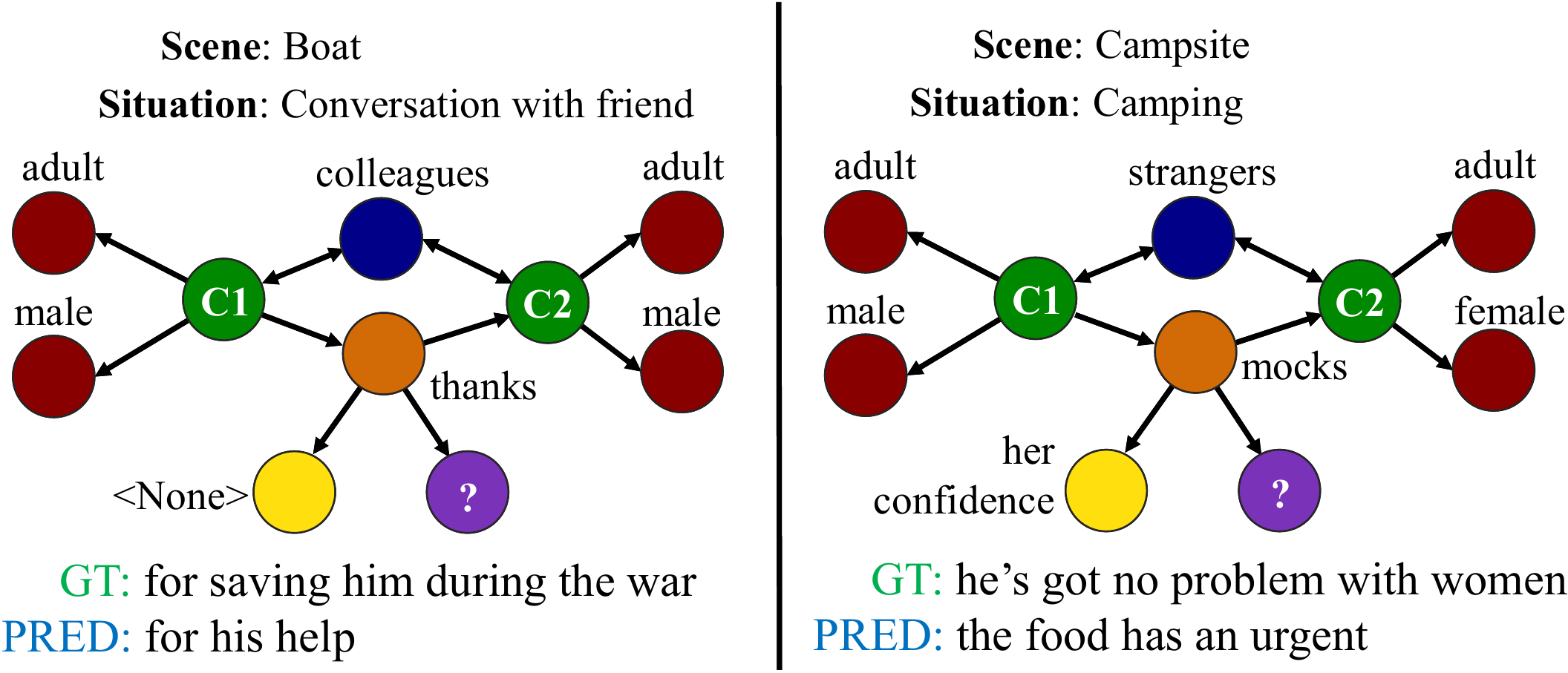}
\vspace{-2mm}
\caption{Example sub-graphs with GT and predicted reasons.
The left one is scored \emph{Very relevant}, and the right \emph{Not relevant}.
However, the model's mistake of relating camping with food is reasonable.
}
\label{fig:reason_predictions}
\vspace{-6mm}
\end{figure}

\vspace{-1mm}
\subsection{Interaction Ordering}
\label{subsec:inter_ordering}
\vspace{-1mm}

We measure ordering performance using two metrics:
(i) the model's accuracy at predicting complete sequences; and
(ii) the length of the longest common subsequence (LCS) between the ground-truth and prediction.
Quantitative results are shown in Table~\ref{table:interactionmetrics}.
Table~\ref{table:interactionsorting} shows qualitative examples of the orderings predicted by our model.
The first two sequences are correctly sorted by our model, while the third is a failure case.
However, even in the failure case, interactions 2, 3, 4, and 5 are in the correct order (longest common subsequence), and the entire sequence is plausible.

\begin{table}[]
\vspace{-1mm}
\small
\centering
\label{my-label}
\begin{tabular}{@{}cccc@{}}
\toprule
\textbf{GT} & \textbf{Pred}   & \textbf{Dir.}   & \textbf{Interaction + [Topic]}       \\
\midrule
1   & 1 & $\rightarrow$ & asks [why she's crying]                        \\
2   & 2 & $\leftarrow$  & explains to [why she is sad]                   \\
3   & 3 & $\rightarrow$ & comforts                                       \\
\midrule
1   & 1 & $\rightarrow$ & waits for [to end the audition]                \\
2   & 2 & $\leftarrow$  & informs [audition went bad]                    \\
3   & 3 & $\rightarrow$ & suggests [they have a drink]                   \\
4   & 4 & $\leftarrow$  & agrees                                         \\
\midrule
1   & 2 & $\rightarrow$ & explains [it's great to know who he wants]     \\
2   & 3 & $\rightarrow$ & advises [to go visit her]                      \\
3   & 1 & $\leftarrow$  & expresses doubt [she may not like him]         \\
4   & 6 & $\rightarrow$ & encourages                                     \\
5   & 4 & $\leftarrow$  & thanks [for the advice]                        \\
6   & 5 & $\leftarrow$  & announces [he is going to her]                 \\
\bottomrule
\end{tabular}
\vspace{-1mm}
\caption{\small
Qualitative results for ordering interactions. Each interaction is shown with its topic in brackets. The interactions are listed in their ground-truth order, and the predicted sequence is shown in the ``Pred'' column, where the numbers represent the order in which the interaction is predicted. The $\rightarrow$ indicates that C1 initiates the interaction with C2, and $\leftarrow$ the reverse.}
\label{table:interactionsorting}
\vspace{-5mm}
\end{table}

\vspace{-1mm}
\subsection{Reason Prediction}
\label{subsec:reas_prediction}
\vspace{-1mm}

An interaction can have several distinct and relevant reasons, making automatic scoring using captioning metrics hard. We ask 10 AMT workers to score 100 sub-graphs and their predicted reasons as: \emph{Very relevant}, \emph{Semi-relevant}, and \emph{Not relevant}.
Fig.~\ref{fig:reason_predictions} shows two examples, along with their GT and predicted reasons.
We are able to obtain a clear verdict (6 or more annotators agree) on 72 sub-graphs:
11 samples are rated very relevant, while 10 more are semi-relevant.

\vspace{-3mm}
\section{Conclusion}
\label{sec:conc}
\vspace{-2mm}

In this work, we focused on understanding human-centric situations in videos.
We introduced the \emph{MovieGraphs} dataset, that contains rich annotations of everyday social situations in the form of graphs. Our graphs capture people's interactions, emotions, and motivations, many of which must be inferred from a combination of visual cues and dialog. We performed various statistical analyses of our dataset and proposed three tasks to benchmark situation understanding: graph-based video retrieval, interaction understanding via ordering, and reason prediction. We proposed models for each of the tasks, that point to their successes and challenges.
\blfootnote{\textbf{Acknowledgments.} Supported by the DARPA Explainable AI (XAI) program, NSERC, MERL, and Comcast. We thank NVIDIA for their donation of GPUs. We thank Relu Patrascu for infrastructure support, and we thank the Upwork annotators.}



\section*{Supplementary Material}
\appendix

We provide additional details about our dataset, models, and results.
This supplementary material is structured as follows:

\begin{itemize} \itemsep -3pt
\item We present an ablative study of TF$\cdot$IDF-based retrieval, to examine the discriminative power of each node type (Sec.~\ref{supp:sec:tfidfablativestudy}).
\item We present details of our interaction ordering model, and more qualitative results (Sec.~\ref{supp:sec:interaction_ordering}).
\item We present details of our reason prediction model, and more qualitative results (Sec.~\ref{supp:sec:reason_pred}).
\item We provide details of person detection, clustering, and identification (Sec.~\ref{supp:sec:personid}).
\item We describe the annotation interface used to collect the data (Sec.~\ref{supp:sec:annot_interface}).
\item We present example annotations for one movie from the MovieGraphs dataset (Sec.~\ref{section:moviegraphsexamples}), as well as additional dataset statistics:
  \begin{itemize}
  \item The distributions of the number of characters, interactions, and relationships per clip;
  \item The top-20 relationships and scenes.
  \end{itemize}
\item We provide additional examples of the MovieGraphs dataset (Sec.~\ref{supp:sec:data_stats}), and show visualizations of:
  \begin{itemize}
  \item The emotional profiles and timelines of the main characters from several movies;
  \item Emotions of characters on both sides of interactions/relationships;
  \item A rooted graph of situations;
  \item Examples of interaction annotations.
  \end{itemize}
\end{itemize}

\section{TF$\cdot$IDF Ablative Study}
\label{supp:sec:tfidfablativestudy}

In this section, we present an ablative study of TF$\cdot$IDF graph $\to$ description and graph $\to$ dialog retrieval. In order to gauge the relative discriminative power of each node type for retrieval, we perform the retrieval experiment considering only information in certain node types.

\begin{table}[t]
\centering
\begin{tabular}{@{}lcccc@{}}
\toprule
\textbf{Node Types}     & \textbf{R@1} & \textbf{R@5} & \textbf{R@10} & \textbf{medR} \\ \midrule
\textbf{All Node Types}                          & 61.6   & 83.8    & 89.8    & 1       \\
\textbf{All $\setminus$ \{ Sc., Sit. \}}         & 60.8   & 82.1    & 88.0    & 1       \\
\textbf{All $\setminus$ \{ Sc., Sit., Char.\}}   & 51.0   & 67.8    & 74.2    & 1       \\
\textbf{Scene}                                   & 3.1    & 9.9     & 12.9    & 460     \\
\textbf{Situation}                               & 5.0    & 13.5    & 18.3    & 382     \\
\textbf{Character}                               & 19.0   & 45.3    & 58.7    & 7       \\
\textbf{Attribute}                               & 3.7    & 9.0     & 12.7    & 393     \\
\textbf{Interaction, Summary}                    & 12.8   & 24.6    & 29.9    & 66      \\
\textbf{Interaction}                             & 6.2    & 12.8    & 16.2    & 327     \\
\textbf{Summary}                                 & 7.2    & 15.8    & 20.8    & 264     \\
\textbf{Relationship}                            & 0.6    & 1.8     & 2.6     & 704     \\
\textbf{Topic}                                   & 35.5   & 51.4    & 57.3    & 5       \\
\textbf{Reason}                                  & 24.5   & 38.0    & 42.4    & 26      \\
\bottomrule
\end{tabular}
\vspace{-1mm}
\caption{TF$\cdot$IDF Graph-Description Retrieval Ablation Study}
\label{table:tfidf-graphdesc}
\end{table}

\begin{table}[t]
\centering
\begin{tabular}{@{}lcccc@{}}
\toprule
\textbf{Node Types}      & \textbf{R@1} & \textbf{R@5} & \textbf{R@10} & \textbf{medR} \\ \midrule
\textbf{All Node Types}                        & 31.8   & 49.8   & 57.2  & 6    \\
\textbf{All $\setminus$ \{ Sc., Sit. \}}       & 31.9   & 49.4   & 57.1  & 6    \\
\textbf{All $\setminus$ \{ Sc., Sit., Char.\}} & 31.5   & 46.2   & 51.5  & 8    \\
\textbf{Scene}                                 & 0.7    & 2.2    & 3.1   & 718  \\
\textbf{Situation}                             & 1.2    & 3.5    & 4.8   & 627  \\
\textbf{Character}                             & 6.6    & 17.4   & 23.2  & 58   \\
\textbf{Attribute}                             & 1.0    & 3.4    & 5.7   & 603  \\
\textbf{Interaction, Summary}                  & 3.3    & 6.5    & 8.4   & 580  \\
\textbf{Interaction}                           & 1.8    & 4.1    & 5.8   & 621  \\
\textbf{Summary}                               & 1.6    & 4.1    & 5.8   & 656  \\
\textbf{Relationship}                          & 0.3    & 1.3    & 2.0   & 765  \\
\textbf{Topic}                                 & 28.7   & 42.6   & 47.9  & 15   \\
\textbf{Reason}                                & 17.4   & 28.7   & 32.9  & 122  \\
\bottomrule
\end{tabular}
\vspace{-1mm}
\caption{TF$\cdot$IDF Graph-Dialog Retrieval Ablation Study}
\label{table:tfidf-graphdialog}
\end{table}

We present results on the test set, evaluated using recall @ $\{1, 5, 10\}$ and median rank. Table~\ref{table:tfidf-graphdesc} shows the results for graph $\to$ description retrieval, and Table~\ref{table:tfidf-graphdialog} shows the graph $\to$ dialog retrieval results. In both cases, we find that character, topic, and reason nodes are the most discriminative for localizing the correct clip. This aligns with the intuition that the most relevant information involves who is in the clip, and what the details (topics and reasons) of their interactions are.


\section{Interaction Ordering Details}
\label{supp:sec:interaction_ordering}

In this section, we present details of our interaction ordering model, which uses an attention-based RNN to select interactions sequentially from an input set to form a plausible order.

\paragraph{Dataset Creation for Interaction Ordering.}
We extract all sequences of interactions that occur between each pair of characters in each clip. We consider only interactions that have time stamps, and sort them first based on start time, and then by end time to break ties.

\paragraph{Interaction Sequence Encoding.}
Each training example represents a sequence of interactions between a pair of characters, and consists of: 1) a situation label; 2) a scene label; 3) the relationship(s) between the two characters; 4) the attributes of each character; and 5) a sequence of $N$ interactions (with associated topics).

Consider two characters $C_1$ and $C_2$, with attributes $\mathcal{V}_{C_1}^{att}$ and $\mathcal{V}_{C_2}^{att}$, respectively. Let the scene and situation labels be $v^{sc}$ and $v^{si}$, respectively, and the relationships between $C_1$ and $C_2$ be $v_{C_1 \rightarrow C_2}^{rel}$ and $v_{C_1 \leftarrow C_2}^{rel}$, respectively.

Each interaction $v_{C_1, C_2}^{int}$ with topic $v_{C_1, C_2}^{top}$ is represented by the concatenation of the corresponding GloVe embeddings, $[ \ba_{C_1, C_2}^{int} ; \ba_{C_1, C_2}^{top} ]$, with an additional digit appended to indicate the direction of the interaction: 1 for $C_1 \rightarrow C_2$, -1 for $C_1 \leftarrow C_2$, and 0 for $C_1 \leftrightarrow C_2$.
We denote the interaction representation at step $i$ in the sequence by $\mathbf{x}_i$, so that a sequence of length $N$ is denoted by $\mathbf{X} = \{ \mathbf{x}_1, \dots, \mathbf{x}_N \}$.

\paragraph{Context Encoding.}
We create a global context vector (passed to the decoder RNN at each time step) that incorporates situation, scene, relationship, and attribute information. We restrict attributes to be of age and gender types, and compute attribute encodings as follows:
\begin{equation}
h_{C_1} = W_{C_1} \sum_{k} \alpha_k \ba_{C_1k}^{att}, \quad
h_{C_2} = W_{C_2} \sum_{l} \alpha_l \ba_{C_2l}^{att}
\end{equation}
where the attention weights $\alpha_k$ and $\alpha_l$ are computed using a two-layer MLP.

We encode the relationships between $C_1$ and $C_2$ as follows:
\begin{equation}
h_{r_1} = W_{r_1} \ba_{C_1 \rightarrow C_2}^{rel}, \quad
h_{r_2} = W_{r_2} \ba_{C_1 \leftarrow C_2}^{rel}
\end{equation}
Bidirectional relationships are treated as separate relationships in both directions, $C_1 \rightarrow C_2$ and $C_1 \leftarrow C_2$.

We encode the scene and situation with linear layers on the corresponding GloVe representations:
\begin{eqnarray}
h_{sc} &=& W_{sc} \ba^{sc} \\
h_{si} &=& W_{si} \ba^{si}
\end{eqnarray}

Finally, we combine the encoded components to form the global context vector:
\begin{equation}
z = h_{C_1} + h_{C_2} + h_{r_1} + h_{r_2} + h_{sc} + h_{si}
\end{equation}

\paragraph{Attention-Based Decoder RNN.}
We use a single-layer Gated Recurrent Unit (GRU) RNN to select items sequentially from an input set of interactions. At each time step $t$, we compute a local context vector by attending to the input elements:
\begin{equation}
c^{(t)} = \sum_{i=1}^N \alpha_i^{(t)} \mathbf{x}_i
\end{equation}
where
\begin{equation}
\alpha^{(t)} = \text{softmax}(s^{(t)})
\end{equation}
and
\begin{equation}
s_i^{(t)} = (h^{(t-1)})^T \mathbf{x}_i \, .
\end{equation}
The input to the GRU at time $t$ consists of 1) an input element $\mathbf{x}^{(t)}$ (described below for the train and test settings); 2) the local context $c^{(t)}$; and 3) the global context $z$:
\begin{equation}
[ \mathbf{x}^{(t)} ; c^{(t)}; z ] \, .
\end{equation}
The output is computed as a linear transformation on the hidden state:
\begin{equation}
o^{(t)} = W_o h^{(t)}
\end{equation}

We score each interaction from the input set by computing the inner product between the output $o^{(t)}$ and each of the interaction representations $\mathbf{x}_i$.
The model is trained end-to-end with cross-entropy loss on these scores.
At test time, the input element with the highest score is chosen at each time step, and is masked out from the input set so that it is not selected again in future steps.
At training time, we use teacher forcing (i.e., choosing the correct ground-truth interaction to be passed forward to the next step, regardless of which interaction scored the highest) 50\% of the time, and the model's own predictions 50\% of the time.

\paragraph{Results.}
In Table~\ref{table:extrainteractionsorting} we show additional qualitative interaction-ordering results.
In examples (a)-(d) the predicted order matches the ground-truth, while in examples (e)-(i) the predicted order does not match. However, even though the orders presented in examples (e)-(i) are not identical to the ground-truth orders, they are still plausible.

\begin{table}[t]
\tabcolsep=0.10cm
\small
\centering
\begin{tabular}{@{}ccccc@{}}
\toprule
\textbf{Ex.} & \textbf{GT} & \textbf{Pred}  &  \textbf{Dir.}  &  \textbf{Interaction + [Topic]}     \\
\midrule
\multirow{ 3}{*}{a)}  & 1   & 1 & $\rightarrow$ & asks [who was on the phone]               \\
                     & 2   & 2 & $\leftarrow$  & informs [a friend called]                 \\
                     & 3   & 3 & $\rightarrow$ & orders [to go care for the child]         \\
\midrule
\multirow{ 3}{*}{b)}  & 1   & 1 & $\rightarrow$ & calls to                                  \\
                     & 2   & 2 & $\leftarrow$  & ignores                                   \\
                     & 3   & 3 & $\rightarrow$ & offers [treat]                            \\
\midrule
\multirow{ 4}{*}{c)}  & 1   & 1 & $\rightarrow$ & asks [will they ever be happy]            \\
                     & 2   & 2 & $\leftarrow$  & answers [that they are happy]             \\
                     & 3   & 3 & $\leftarrow$  & suggests [they borrow her old crib]       \\
                     & 4   & 4 & $\rightarrow$ & hushes                                    \\
\midrule
\multirow{ 4}{*}{d)}  & 1   & 1 & $\rightarrow$ & asks about [identity]                            \\
                     & 2   & 2 & $\leftarrow$  & yells at                                         \\
                     & 3   & 3 & $\rightarrow$ & yells [to get out]                               \\
                     & 4   & 4 & $\leftarrow$  & accuses [that he pretends to be a millionaire]   \\
\midrule
\multirow{ 5}{*}{e)}  & 1   & 4 & $\rightarrow$ & confesses to [he did not murder anyone]          \\
                     & 2   & 5 & $\rightarrow$ & explains to [he went to jail for video pirating] \\
                     & 3   & 3 & $\leftarrow$  & reproaches [they thought he was a murderer]      \\
                     & 4   & 1 & $\leftarrow$  & asks [about his gun]                             \\
                     & 5   & 2 & $\rightarrow$ & shows [he doesn't have a gun]                    \\
\midrule
\multirow{ 5}{*}{f)}  & 1   & 4 & $\rightarrow$ & begs [not to go]                                 \\
                     & 2   & 2 & $\leftarrow$  & apologizes                                       \\
                     & 3   & 5 & $\leftarrow$  & explains to [she's retiring]                     \\
                     & 4   & 3 & $\leftarrow$  & encourages [to be strong]                        \\
                     & 5   & 1 & $\rightarrow$ & calls after                                      \\
\midrule
\multirow{ 5}{*}{g)}  & 1   & 1 & $\rightarrow$ & orders [to shut up]                              \\
                     & 2   & 2 & $\leftarrow$  & accuses [of assault and battery]                 \\
                     & 3   & 4 & $\rightarrow$ & threatens [to beat him up]                       \\
                     & 4   & 5 & $\rightarrow$ & warns [he must make it up to his friend]         \\
                     & 5   & 3 & $\rightarrow$ & admits [he doesn't like to yell]                 \\
\midrule
\multirow{ 5}{*}{h)}  & 1   & 1 & $\rightarrow$ & informs [they lost all the stuff]                \\
                     & 2   & 4 & $\leftarrow$  & asks [if ex-worker set the fire]                 \\
                     & 3   & 3 & $\rightarrow$ & explains to [the cause of her husband's death]   \\
                     & 4   & 2 & $\leftarrow$  & informs [that they have to succeed]              \\
                     & 5   & 5 & $\rightarrow$ & reassures [they won't let them win]              \\
\midrule
\multirow{ 5}{*}{i)}  & 1   & 3 & $\rightarrow$ & asks [why her eyes look old]                          \\
                     & 2   & 4 & $\leftarrow$  & jokes [his eyes look kind]                            \\
                     & 3   & 2 & $\leftarrow$  & explains to [she woke up early]                       \\
                     & 4   & 1 & $\rightarrow$ & interrupts [to stop her from clearing his plate]      \\
\bottomrule
\end{tabular}
\caption{Additional qualitative results for interaction ordering. The GT column shows the ground-truth order of interactions, the Pred column shows the order in which the interactions were chosen by the model, and the Dir column shows the direction of the interaction between the characters.}
\label{table:extrainteractionsorting}
\end{table}

\section{Reason Prediction Details}
\label{supp:sec:reason_pred}

In this section, we present details of our reason prediction model.
In particular, we build a model that encodes the context of the interaction and generates plausible reasons using a decoder RNN.

\paragraph{Context Encoding.}
The context is comprised of:
(i) a pair of characters $C_1, C_2$, represented by their attributes $\mathcal{V}^{att}_{C_1}$ and $\mathcal{V}^{att}_{C_2}$ (actual names don't tell us anything and do not matter);
(ii) the interaction $v^{int}_{C_1,C_2}$ for which we wish to predict the reason;
(iii) the relationship $v^{rel}_{C_1,C_2}$ between the characters;
(iv) the scene $v^{sc}$ and situation $v^{si}$ labels;
(v) and optionally, a topic $v^{top}_{C_1,C_2}$ node associated with the interaction.

In particular, we restrict attributes to age and gender (as others, such as emotions, were found to have little influence).
The characters are represented by a weighted combination of attributes:
\begin{equation}
h_{C_1} = \sum_k \alpha_k \ba^{att}_{C_1k}, \quad
h_{C_2} = \sum_l \alpha_l \ba^{att}_{C_2l}
\end{equation}
where the attention weights $\alpha_k, \alpha_l$ are learned using a two-layer MLP.
$\ba_{\cdot}$ corresponds to the 100-d GloVe embedding for each node.

In our graphs, interactions and relationships can take three possible directions:
$C_1 \rightarrow C_2$ (\eg~parent), $C_1 \leftarrow C_2$ (\eg~child), and $C_1 \leftrightarrow C_2$ (\eg~friends).
We first encode a direction $C_e \rightarrow C_f$:
\begin{equation}
h_d^{e \rightarrow f} = W_{C_d} \cdot h_{C_e} + W_{C_r} \cdot h_{C_f} \, ,
\end{equation}
where $W_{C_d}$ corresponds to the parameters for the \emph{doer} and $W_{C_r}$ the \emph{receiver}.
Interactions and relationships are encoded as:
\begin{eqnarray}
h_i^{e \rightarrow f} &=& h_d^{e \rightarrow f} + W_i \cdot (\ba^{int}_{C_1,C_2} + \ba^{top}_{C_1,C_2}) \, ,\\
h_r^{e \rightarrow f} &=& h_d^{e \rightarrow f} + W_r \cdot \ba^{rel}_{C_1,C_2} \, ,\\
h_{i|r}^{e \leftrightarrow f} &=& (h_{i|r}^{e \rightarrow f} + h_{i|r}^{e \leftarrow f}) / 2 \, .
\end{eqnarray}

The final context vector is a linear combination:
\begin{equation}
h_c = W_p \ba^{sc} + W_s \ba^{si} + h_i + h_r \, .
\end{equation}

\begin{figure*}[t]
\vspace{-0mm}
\centering
    \includegraphics[width=\linewidth]{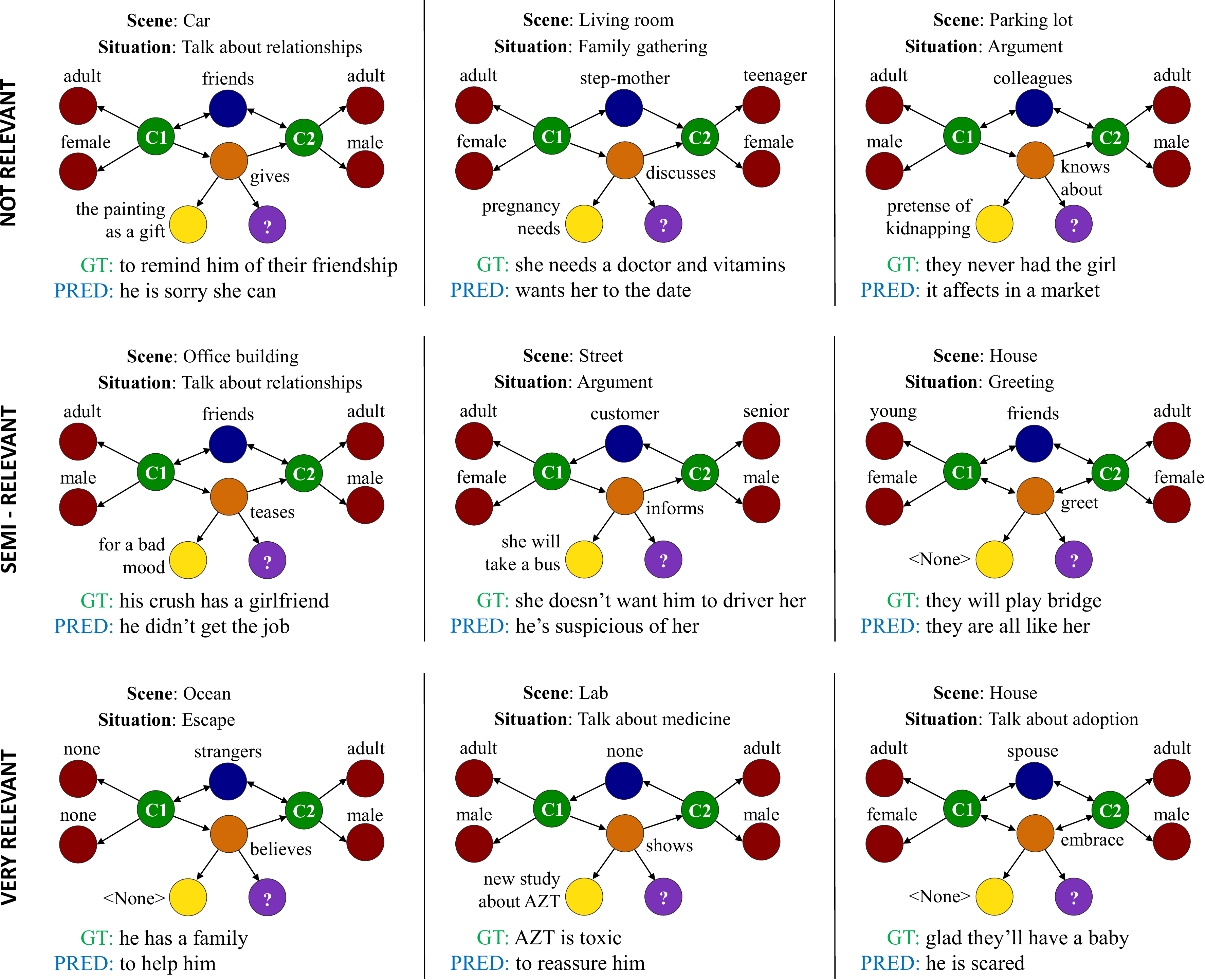}
    \vspace{-3mm}
\caption{Reason prediction results on the test set, grouped by the three evaluation labels.}
\label{fig:reason_pred_results}
\vspace{-3mm}
\end{figure*}

\paragraph{Decoder RNN.}
We adopt a Gated Recurrent Unit RNN~\cite{cho2014GRU} with a 100-d hidden state as our decoder.
The context is fed in at each time-step along with the previous sampled word (or \texttt{START} token).
We use a temperature sampling scheme to generate variability in the predicted reasons.

\paragraph{Results.}

Fig.~\ref{fig:reason_pred_results} shows several more examples of context graphs and ground-truth and predicted reasons.
Each row shows the results for test set samples annotated as \emph{Not relevant} (row 1), \emph{Semi-relevant} (row 2) and \emph{Very relevant} (row 3).
Note that the results in the last row, while not predicting the ground-truth reason, are very plausible, affirming the difficulty of this task.

\section{Person Detection, Clustering, and Identification}
\label{supp:sec:personid}

\paragraph{Data Collection for Face ID.}
We grounded each character in the graph with all the face tracks in the clip (obtained with OpenFace~\cite{baltruvsaitis2016openface}).
As there are on average 1740 face tracks per movie, 9.6 per clip, and over 88K in our full dataset (prior to removing false positives and track switches), annotating the tracks directly would have been time consuming.
Instead, we first linked face tracks that belong to the same person in a clip by performing hierarchical agglomerative clustering on upper body color histogram features (a character wears the same clothing during a clip) corresponding to the face tracks.
The annotator is then asked to assign character names (from those found in the graph) to clusters in the clip. In cases where clustering is wrong, we break the cluster into face tracks, and the annotator labels each track.

\vspace{-2mm}
\paragraph{Face Detection and Tracking.}
Face tracks are extracted using the OpenFace tracker \cite{baltruvsaitis2016openface}.
The output of the tracker are face detections in each frame, which we need to post-process into face tracks.
To match detections from one frame to the detections in subsequent frames, we compute the IoU (intersection over union) between each possible pair of detections from the different frames.
We then compute the optimal matching of pairs that maximizes the IoU between pairs using the Hungarian algorithm.
Since detections might be spurious, we allow for gaps of up to 5 frames in which the face might not be detected in the same face track and we discard every face track that lasts less than 10 frames.
We note that the quality of our face tracker is primarily limited by the face detector, which occasionally results in missing face tracks and false positives.
Furthermore, while we found our post-processing heuristics to work reasonably in practice, there are some track switches grouped together.
On average, we obtain 66 false positive face tracks and track switches per movie (corresponding to less than 4\% of all tracks).

\vspace{-2mm}
\paragraph{Face Features.}
We compute face embeddings for the purpose of character identification and face track clustering.
Our model builds upon the VGG-Face model of~\cite{parkhi2015deep}.
We extend it by (i) reducing the dimensionality of the features from 4096 to 128 with a linear projection layer (\ie, a fully connected layer without non-linearity),
and (ii) making them have a unit norm.
We initialize our model with the weights of~\cite{parkhi2015deep} and train it using a triplet loss as described in~\cite{schroff2015facenet} with a margin $m = 0.2$ but without mining hard negatives due to limited computational resources.

To train our model we use our own dataset, comprised of images from face tracks and cast pictures from IMDb.
Face track identities are annotated in the dataset, while IMDb pictures have associated identity metadata.
Furthermore, we filter IMDb pictures by running a standard face detection algorithm~\cite{viola2001rapid} and including only those with a single face detection.
Each triplet in our dataset is formed by a face track anchor, a positive example randomly sampled either from matching face tracks or IMDb pictures, and a negative example also from either non-matching IMDb pictures or face tracks.
Since a face track contains many images, we randomly select one of them to represent a face track example in a triplet during training.
At test time, a face track is represented by the average features of its images.
We feed face crops into our model instead of full images, expanding the face detection bounding box by a factor of 2 to also include clothing and hair.
The context was found to be especially helpful for track clustering.
Expanded crops are resized to a fixed resolution of 224x224 using bilinear interpolation.

\vspace{-2mm}
\paragraph{Face Track Clustering.}
We perform face track clustering in each movie scene to group together face tracks belonging to the same characters.
To perform clustering, we use face features for each of the face tracks and compute pairwise distances between them.
We then apply hierarchical agglomerative clustering with a cut-off at threshold $t= 0.75$, empirically chosen to minimize the OCI metric on the validation set.

We evaluate our clustering performance in the test set with three different metrics:
(i) cluster purity, indicating how many clusters capture a single identity,
(ii) weighted cluster purity, in which the purity of a cluster is weighted by the number of face tracks in the cluster, and
(iii) operator clicks index (OCI) \cite{Guillaumin2009}, indicating the number of face tracks in wrong clusters plus the total number of clusters obtained.
This last metric seeks to evaluate the cost of annotating face tracks given a clustering.
While evaluating our method, we discard wrong face tracks or face tracks that switch faces.
The performance of our method is shown in Table~\ref{table:clustering_performance}.

\begin{table}[h]
    \centering
    \begin{tabular}{cc}
        \toprule
        \textbf{Metric}      & \textbf{Score}  \\ \midrule
        \textbf{Cluster Purity}   & 75.7\% \\
        \textbf{Weighted Purity}  & 75.8\% \\
        \textbf{OCI}              & 6.4 \\
        \bottomrule
    \end{tabular}
    \caption{Clustering performance}
    \label{table:clustering_performance}
\end{table}

\vspace{-4mm}
\paragraph{Character Identification.}
Here we explore the task of assigning character identities to face tracks.
We extract face features for cast pictures with a single face detection and compute pairwise distances with face track images.
We then rank the images by distance to the face tracks and assign character identity probabilities to each face track.
We evaluate our assignments with accuracy in Table~\ref{table:character_id_performance}.
We keep the top-15 characters that appear in at least 10 graphs, as otherwise, movies (including all extras and background) can have characters (and actors) that often do not even have pictures on IMDb.

\begin{table}[h]
    \centering
    \begin{tabular}{ccc}
        \toprule
        \textbf{Metric}      & \textbf{Ours} & \textbf{Random}  \\ \midrule
        \textbf{Accuracy}    & 43.7\%          & 13.2\% \\
        \bottomrule
    \end{tabular}
    \caption{Character identification}
    \label{table:character_id_performance}
\end{table}

\begin{figure*}[t]
\centering
   \includegraphics[width=\linewidth]{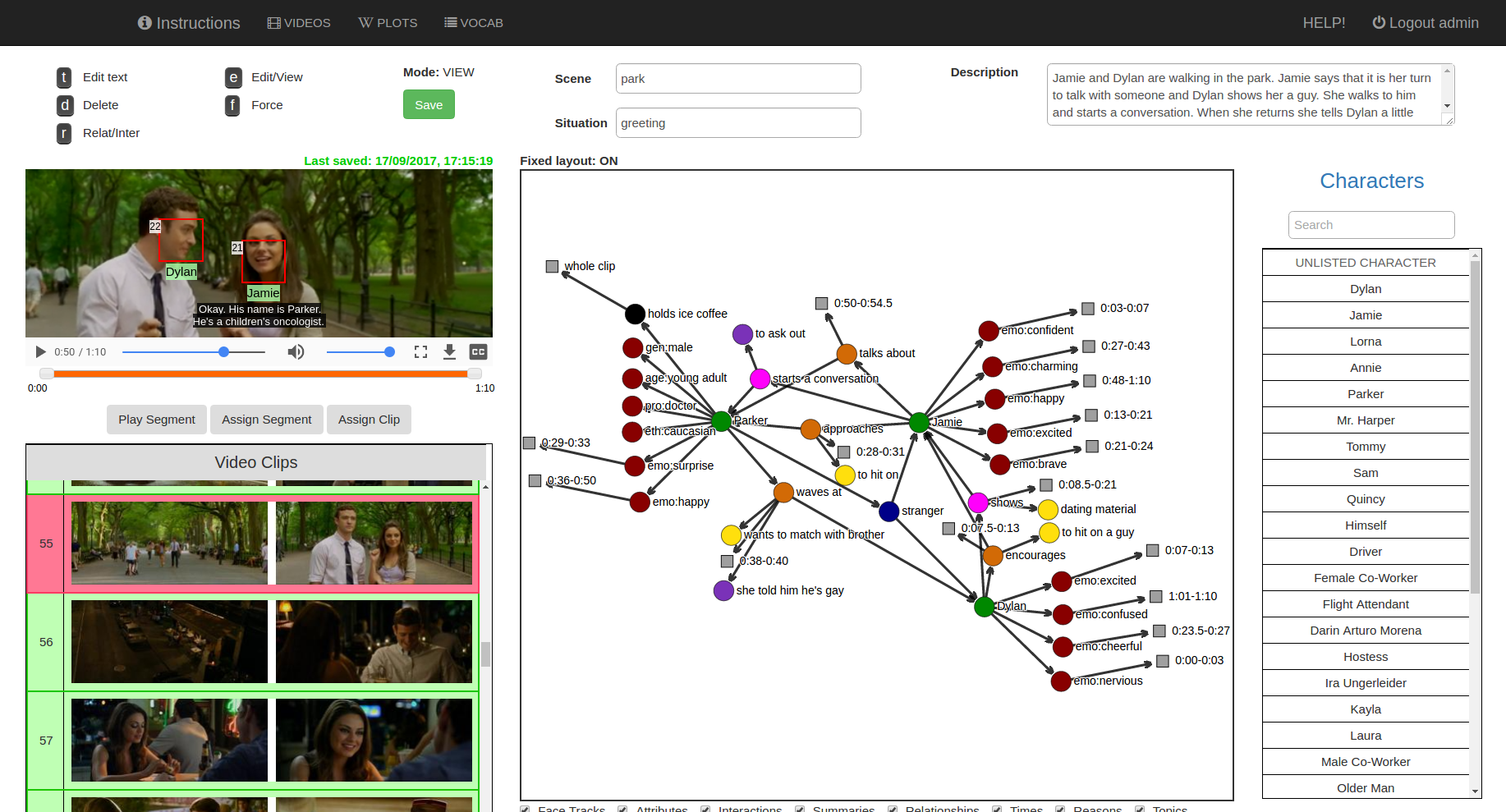}
\caption{\small The MovieGraphs annotation interface provides four separate areas where annotators type the scene label, the situation label, and the natural language description, and where they create the graph. On the left, they select a clip from a sequence of clips, and on the right, they select characters from a cast list (obtained from IMDb).}
\label{fig:interface}
\end{figure*}

\section{Annotation Interface}
\label{supp:sec:annot_interface}

In this section, we describe our data collection procedure.
We developed a custom web-based annotation interface (Fig.~\ref{fig:interface}) to support our data collection process. An annotator first selects a movie clip from the list on the left-hand side; the clip to be annotated then plays at the top. For each clip, the annotator must annotate four elements: 1) a scene label; 2) a situation label; 3) a natural language description; and 4) a graph. The graph canvas is in the center of the page; text boxes for scene and situation labels, as well as for the description, are above. In order to create a graph, an annotator starts by adding \textit{characters}. The list of characters (from the IMDb cast of the movie) is shown on the right; clicking a character name from the list creates the corresponding character node in the graph canvas. An unlisted character can be added by choosing \textit{unlisted character} and then naming it. All other node types (e.g., attributes, interactions, relationships, topics, reasons, and time stamps) are added directly in the graph canvas. We provided initial vocabularies for scenes and situations which annotators could select from, but they were also allowed to add custom items to the lists. The interactions, topics, and reasons were free-form, and we encouraged the annotators to be concise.

\paragraph{Workflow.}
Each annotator watched a movie from start to finish, annotating each clip in order.
This ensured that each annotation contained data inferred from the movie up until that point in time, and allowed annotators to gain insight into the reasons behind each character's actions.
The typical workflow of the annotator was to watch the clip and first write the description in natural language, then provide the scene and situation label, and finally create the graph.
Good data requires proper identification of the scene and situation, a description thorough enough that people reading it can understand what happens in the clip, and a graph detailed enough that reading it conveys the gist of the clip, as well.

\vspace{-3mm}
\paragraph{Training Annotators on Upwork.} We hired annotators through the freelance service Upwork. Therefore, our setup had to overcome some unique challenges, the main one being uniformity: the annotators have to agree on the level of detail they put in the graph. To support data quality, we trained annotators on a common set of clips, and continued monitoring the annotations.

\vspace{-3mm}
\paragraph{Dividing Movies into Clips.}  To obtain the clips for annotation, we automatically split each movie into scenes ~\cite{Tapaswi2014_StoryGraphs} and then grouped the scenes into clips manually, such that each clip corresponds to one coherent social situation (e.g., \textit{party}). As some situations were longer than others, our clips vary in length from around 20 seconds to 2 minutes, with an average length of 44 seconds.

\section{MovieGraphs Examples and Statistics}
\label{section:moviegraphsexamples}

\begin{figure*}[p]
\centering
   \includegraphics[width=\linewidth]{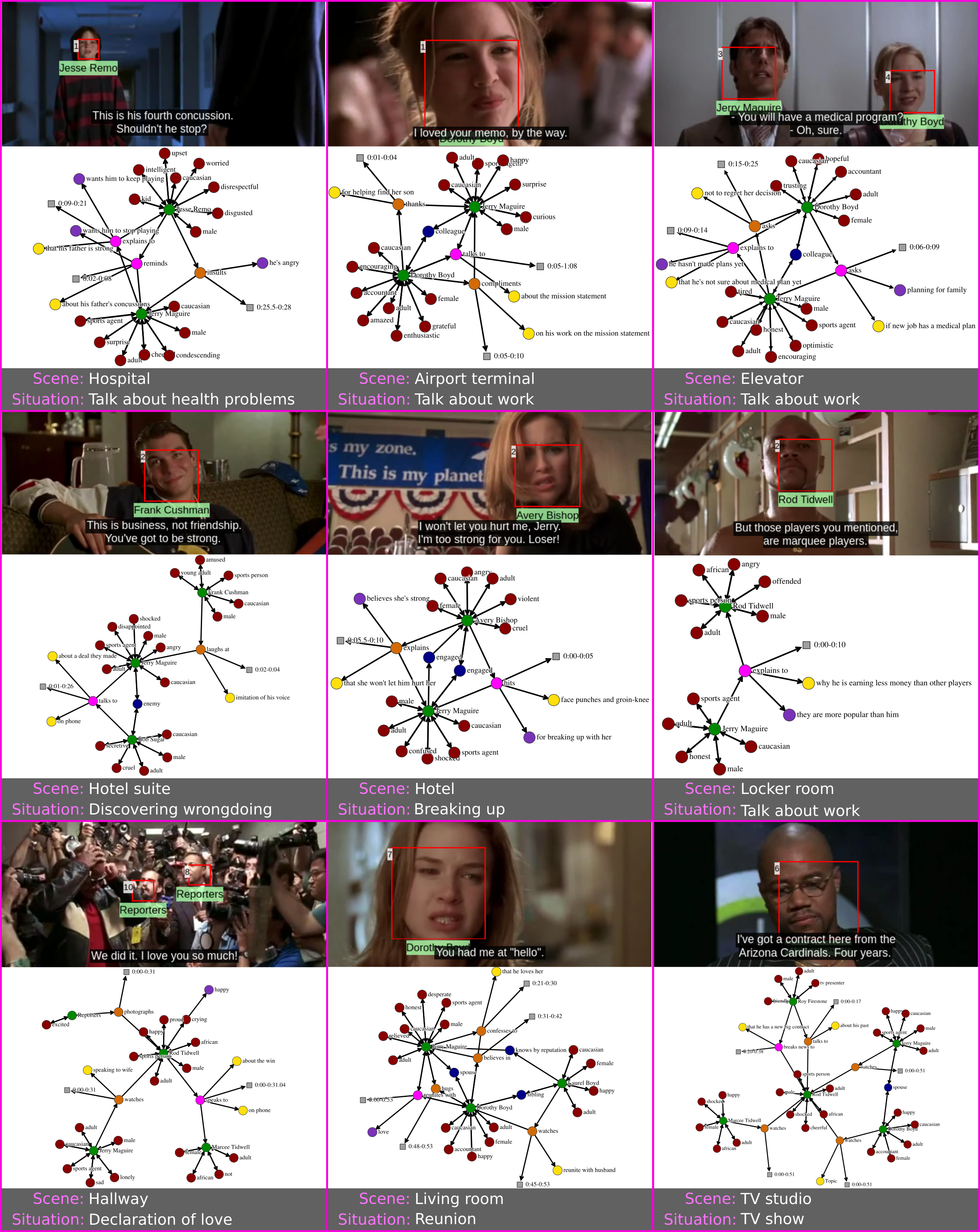}
\caption{\small Example annotations of various clips throughout the movie ``Jerry Maguire,'' showing scenes, situations, and graphs. Each clip is also annotated with a natural language description (not shown due to space constraints).}
\label{fig:jerrymaguirespread}
\end{figure*}

We show examples of the graph annotations for the movie ``Jerry Maguire'' in Fig.~\ref{fig:jerrymaguirespread}. We sample nine graphs from various points in the movie, to show the progression of the story.

\begin{figure*}[t]
\centering
    \includegraphics[height=6.7cm]{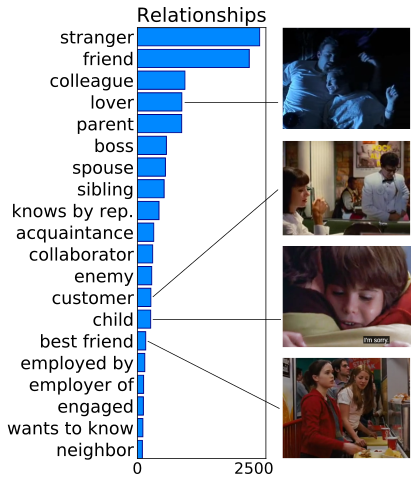} \hspace{6mm}
    \includegraphics[height=6.7cm]{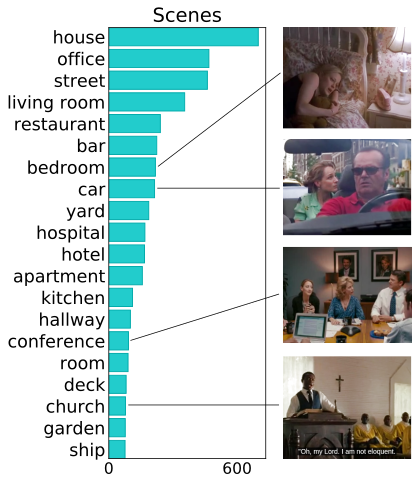}
\caption{\small Distributions of the top 20 relationships and scenes.}
\label{fig:top20relationshipsandscenes}
\end{figure*}

\begin{figure}[]
\centering
    \includegraphics[width=\linewidth]{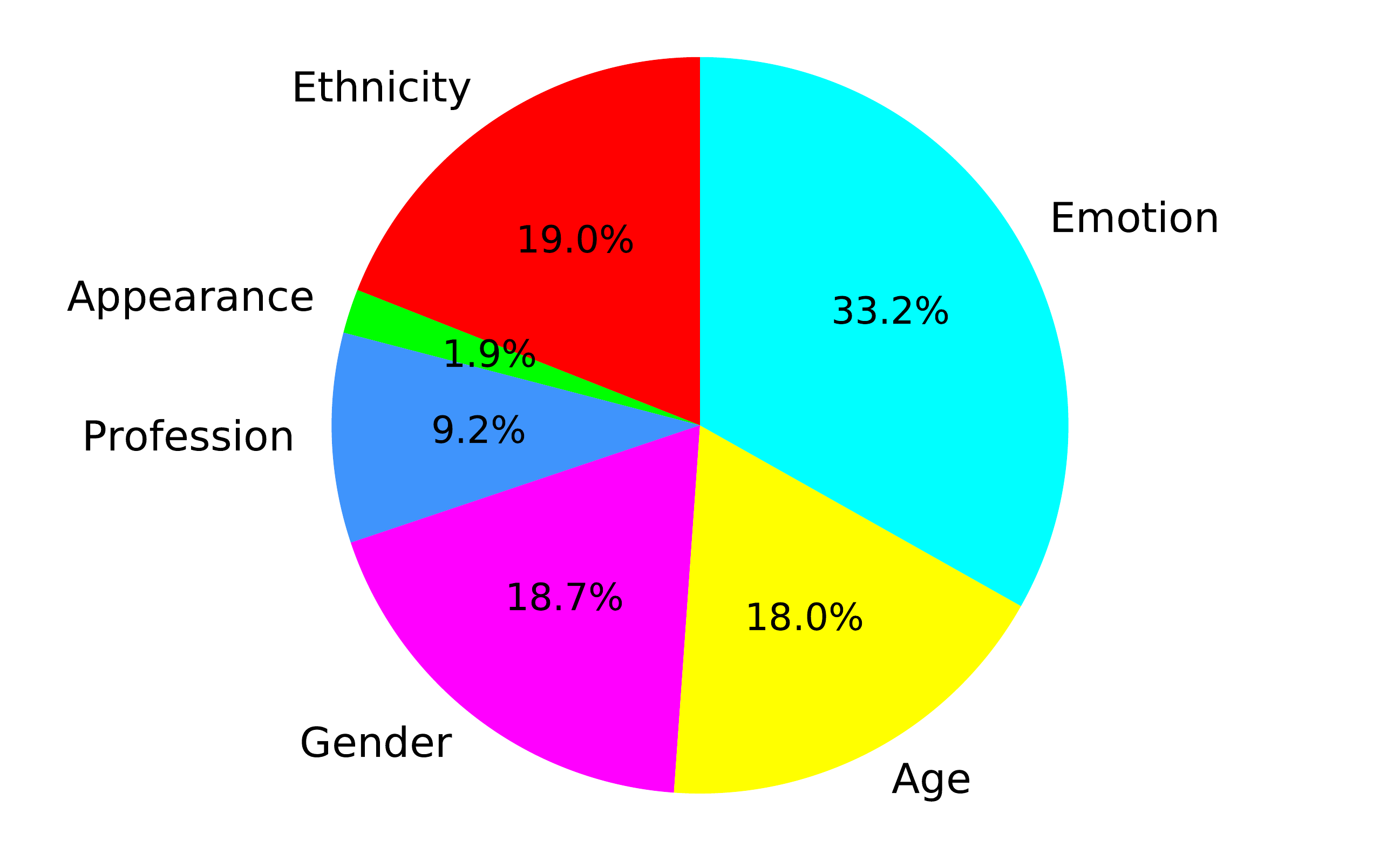}
\caption{\small The distribution of attribute types.}
\label{fig:attributepiechart}
\end{figure}

\begin{figure}[]
\centering
    \includegraphics[width=0.9\linewidth]{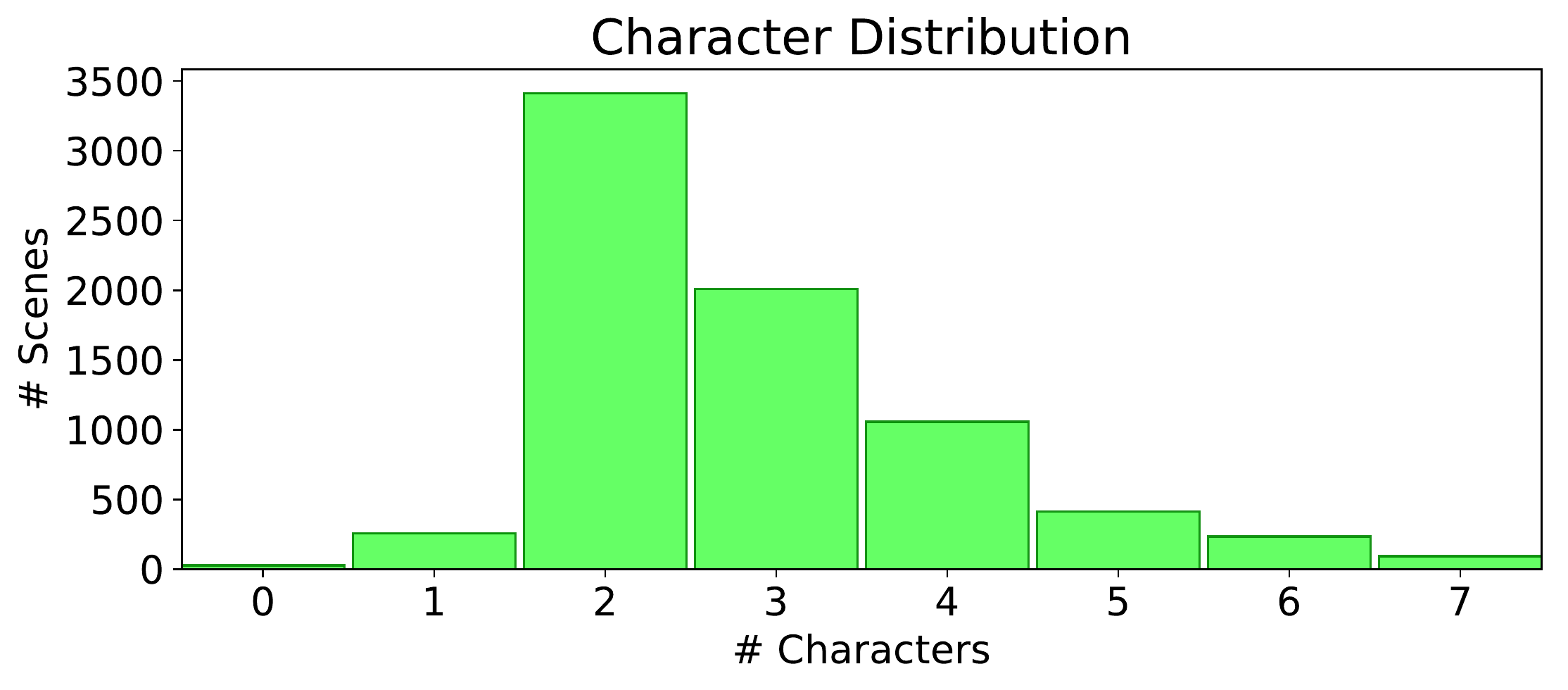}
\caption{\small The distribution of the number of characters per clip, over all movies.}
\label{fig:scenecharacterdistribution}
\end{figure}

\begin{figure}[]
\centering
    \includegraphics[width=0.9\linewidth]{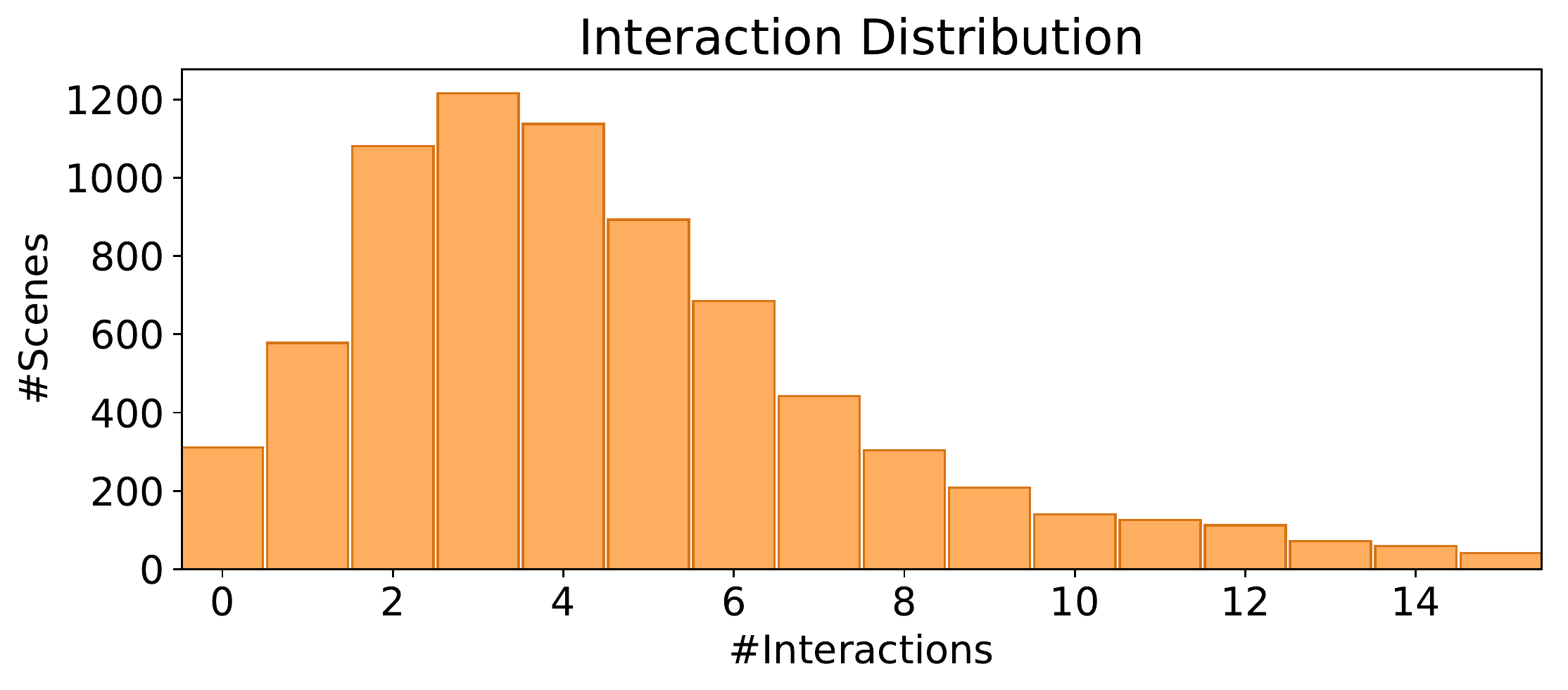}
\caption{The distribution of the number of interactions per clip, over all movies.}
\label{fig:interactiondistribution}
\end{figure}

\begin{figure}[]
\centering
    \includegraphics[width=\linewidth]{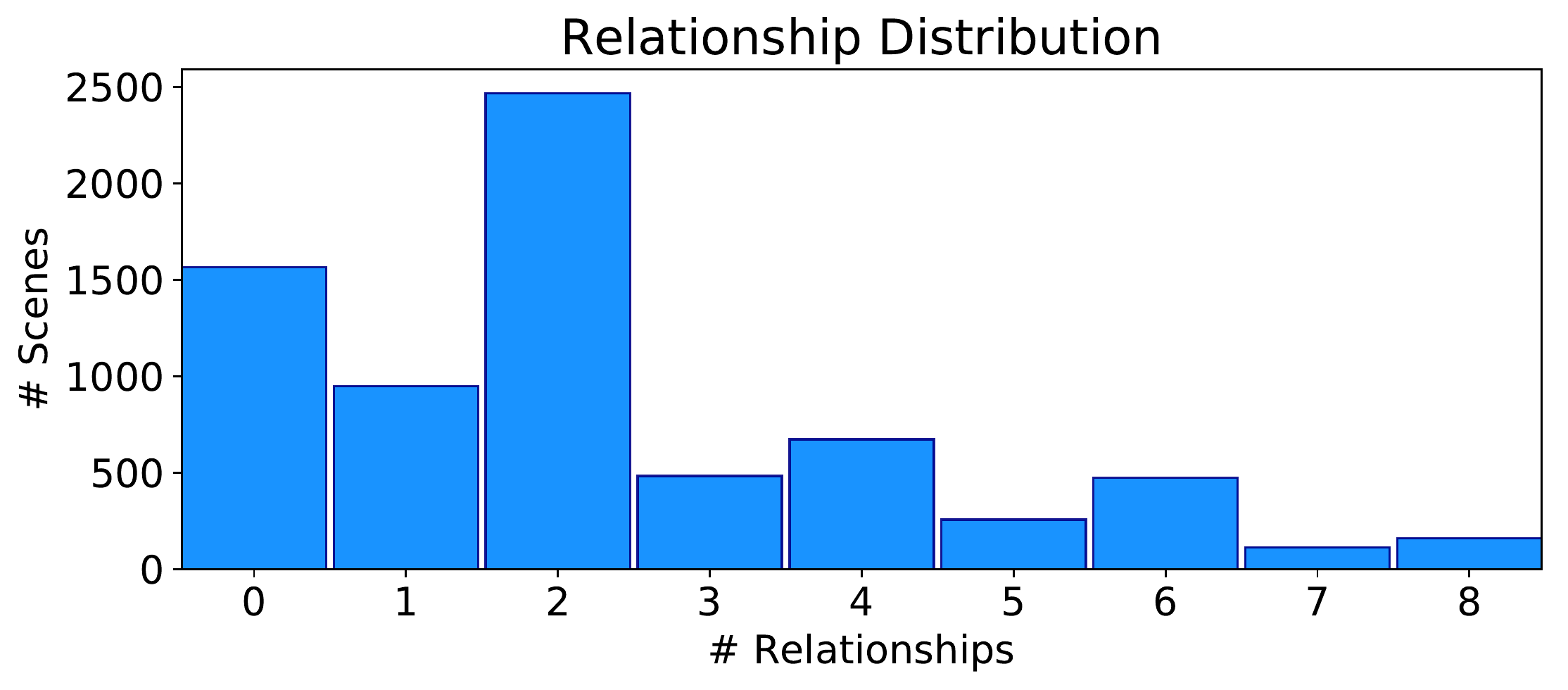}
\caption{The distribution of the number of relationships per clip, over all movies.}
\label{fig:relationshipdistribution}
\end{figure}

Fig.~\ref{fig:top20relationshipsandscenes} shows the top 20 relationships and scenes across all movies.
The distribution of attribute types is shown in Fig.~\ref{fig:attributepiechart}.
We also present the distributions of the number of nodes of different types in each clip:
the number of characters per clip is shown in Fig.~\ref{fig:scenecharacterdistribution}; the
number of interactions in Fig.~\ref{fig:interactiondistribution}; and the
number of relationships in Fig.~\ref{fig:relationshipdistribution}.

\clearpage

\section{More Examples}
\label{supp:sec:data_stats}

\paragraph{Character Emotional Profiles.}

\begin{figure*}[h]
\centering
    \includegraphics[width=0.47\linewidth]{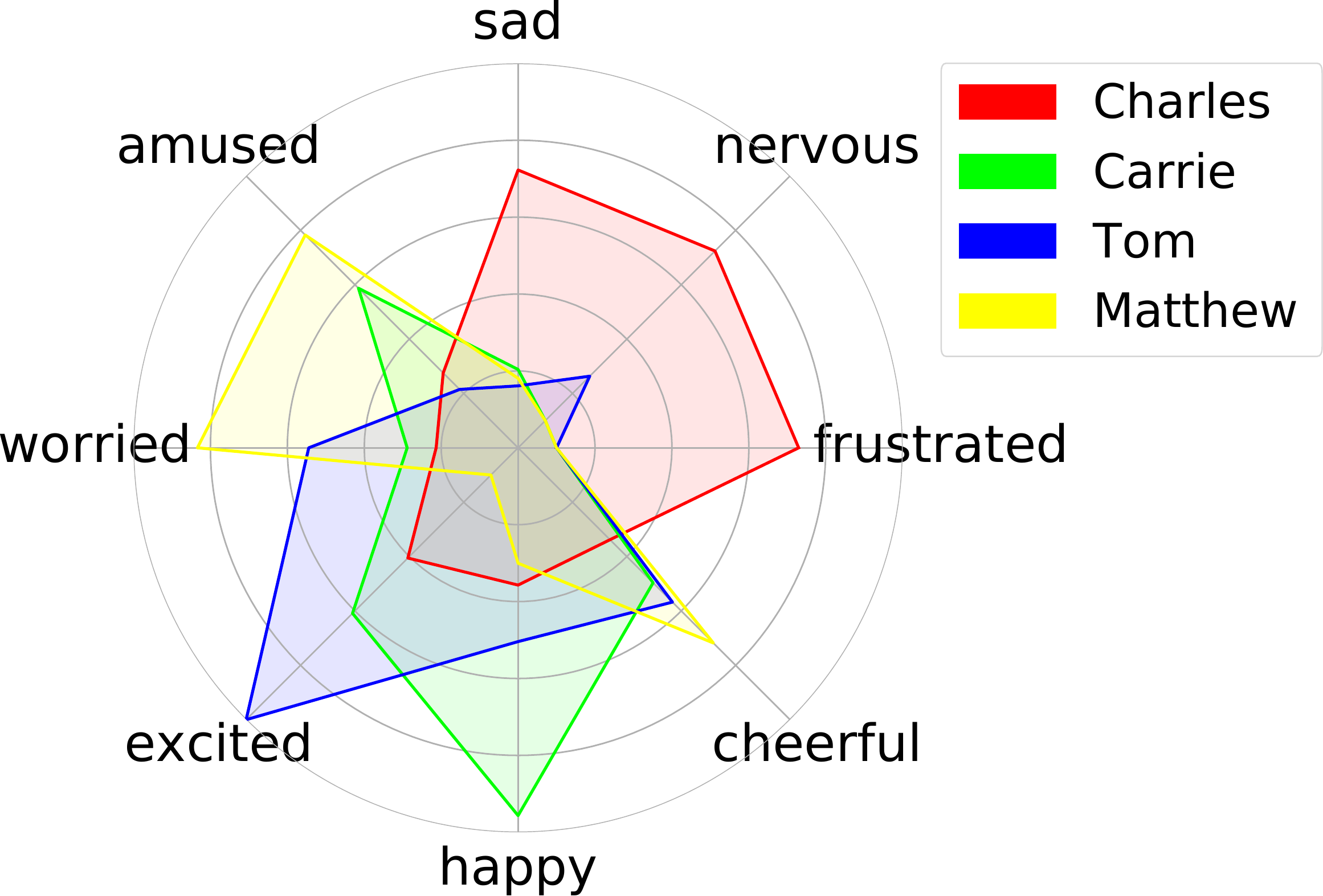}
    \includegraphics[width=0.49\linewidth]{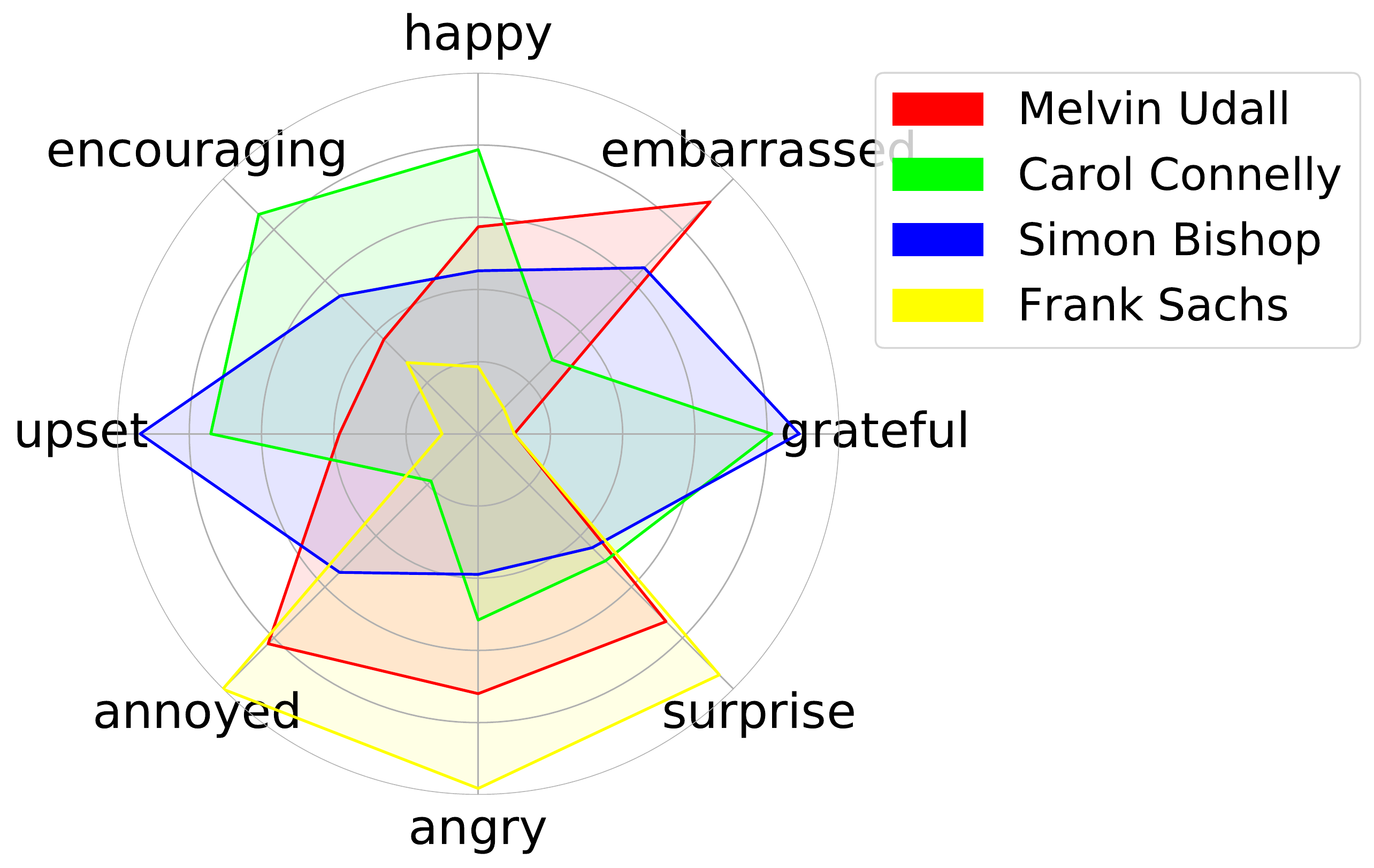}
\caption{\small Emotional profiles of the main characters in ``Four Weddings and a Funeral'' (left) and ``As Good as It Gets'' (right). These profiles were obtained by aggregating the emotions of these characters over the whole movie.}
\label{fig:radarplot_fourweddings_asgood}
\end{figure*}

We show the emotional profiles of the main characters in several movies, based on the emotions annotated for each character across all the clips in a movie. We see that in the movie ``Four Weddings and a Funeral'' (Fig.~\ref{fig:radarplot_fourweddings_asgood}, left), characters are mostly sad, nervous (Charles), happy (Carrie), worried (Matthew) and excited (Tom), while in the movie ``As Good As It Gets'' (Fig.~\ref{fig:radarplot_fourweddings_asgood}, right), the characters are often upset or grateful (Simon), embarrassed or angry (Melvin), happy (Carol), and angry or surprised (Frank).

\begin{figure*}[]
\centering
    \includegraphics[width=0.62\linewidth]{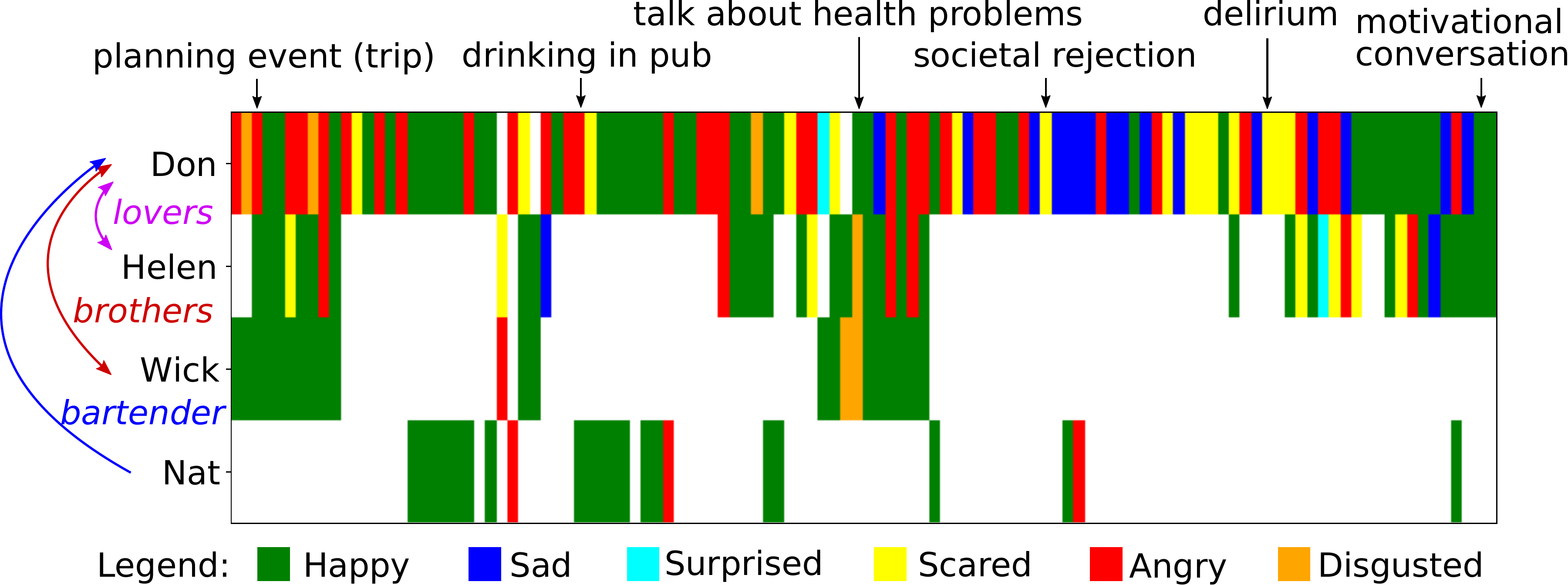} \\
    \vspace{4mm}
    \includegraphics[width=0.62\linewidth]{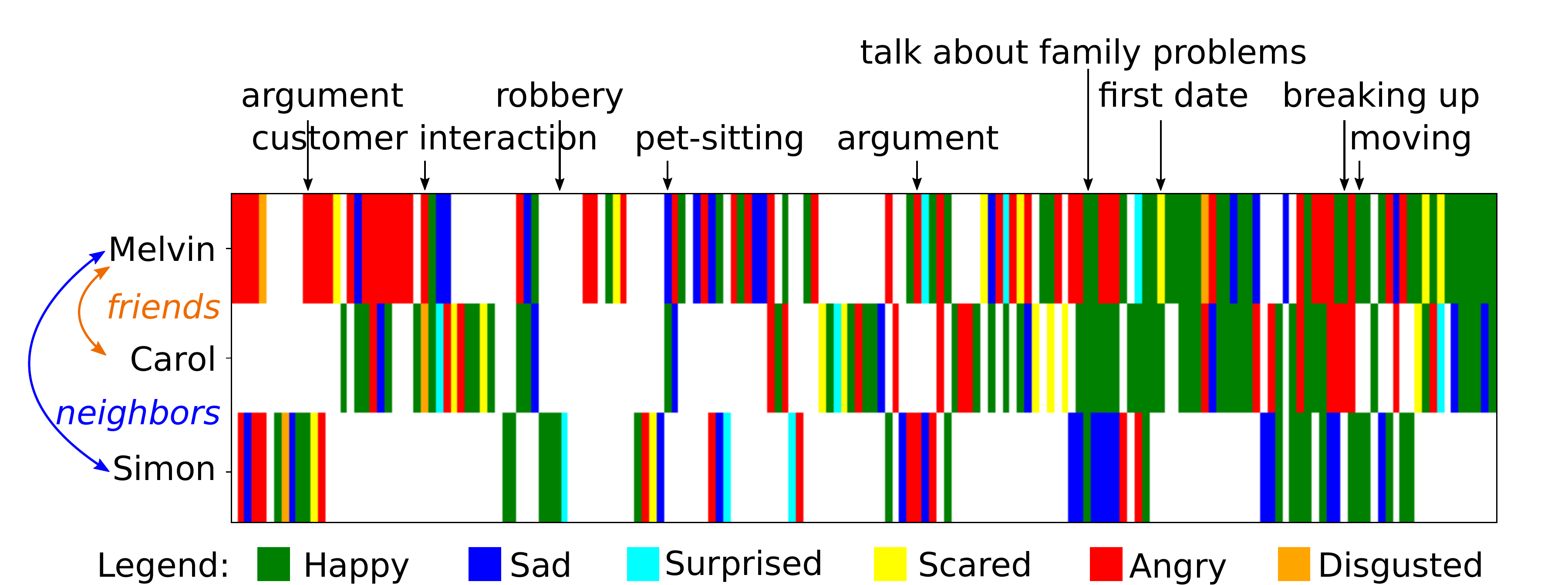} \\
    \vspace{10mm}
    \includegraphics[width=0.62\linewidth]{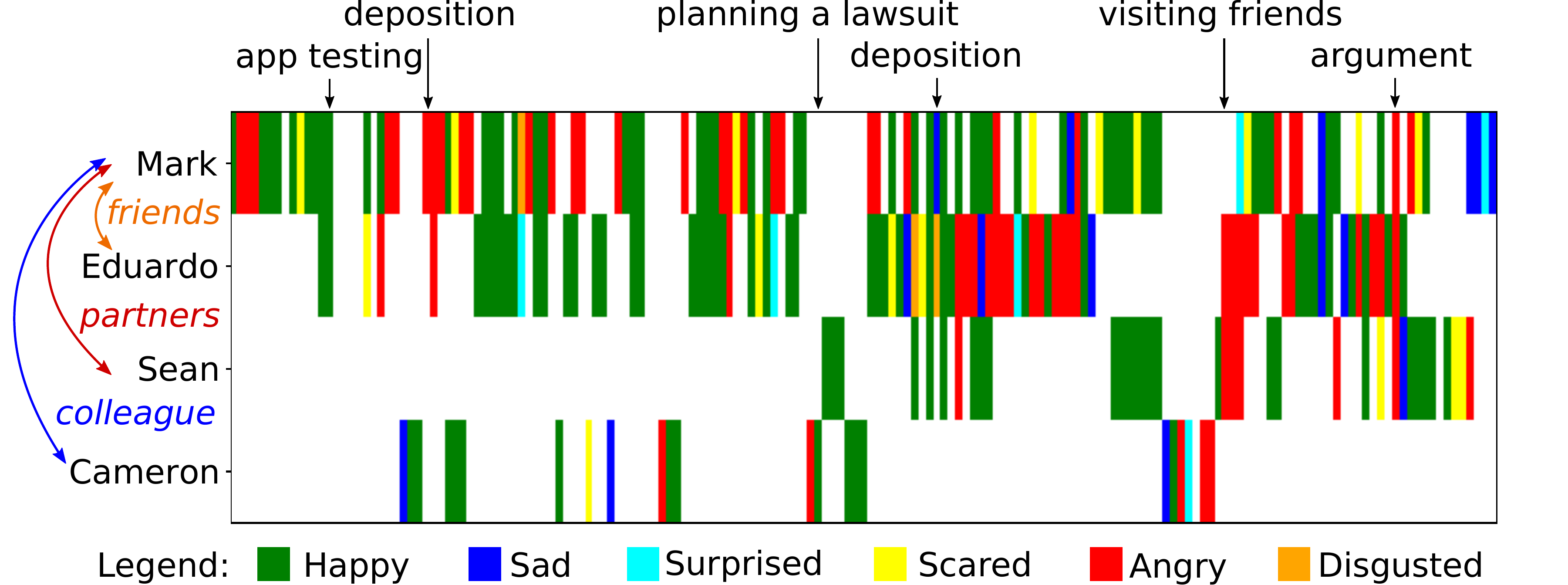} \\
\caption{\small Emotional timelines for the main characters in ``The Lost Weekend'' (top), ``As Good as It Gets'' (middle), and ``The Social Network'' (bottom). The emotional timelines are correlated with situations (shown with arrows above each diagram) and relationships between characters (shown with arrows between names on the left). Wherever a character does not appear in a clip, the space is white.}
\label{fig:emotiontimelines}
\end{figure*}

\paragraph{Emotion Timelines.}
As in real life, characters in movies experience many emotions whose progression during the story we show in Fig.~\ref{fig:emotiontimelines}. We used the six basic emotions (angry, happy, sad, scared, surprised, and disgusted) described in~\cite{ekman1992argument}. We mapped the various words used to annotate emotions in the graphs to these six emotions based on~\cite{feelingvocabulary}, with some manual additions. Where a character was annotated to have several emotions in the same clip, we took the mode of these emotions. The emotions are color-coded, as shown in the legend for our figures.

We then correlated the emotional progression of the main characters in each movie with situations and relationships.
In Fig.~\ref{fig:emotiontimelines}, we show the progression of emotions of the main characters in the movies ``The Lost Weekend,'' ``As Good as It Gets'', and ``The Social Network.''
For example, Fig.~\ref{fig:emotiontimelines} (top diagram) shows the emotional timeline of four characters from the movie ``The Lost Weekend.'' We show that Don, his brother Wick, and Don's girlfriend Helen plan a weekend trip to help Don break his alcohol addiction, but he is angry because he doesn't want to go. Don, still angry, goes to drink in a pub, then becomes sad when he is rejected by society. He continues drinking, and becomes scared during his alcoholic delirium episode. Eventually, he and Helen engage in a motivational conversation, and both end up happy. The timeline shows that Don is a troubled, angry character, while Helen, Wick, and Nat (the bartender) are calming influences.

\begin{figure*}[h]
\centering
    \includegraphics[width=0.67\linewidth]{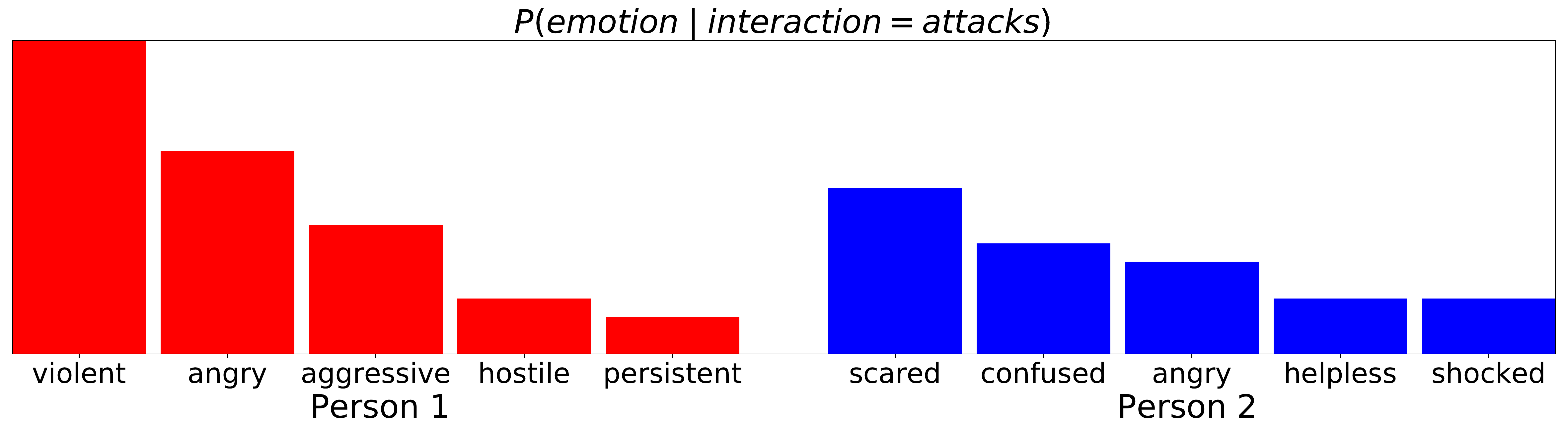} \\
    \vspace{1mm}
    \includegraphics[width=0.67\linewidth]{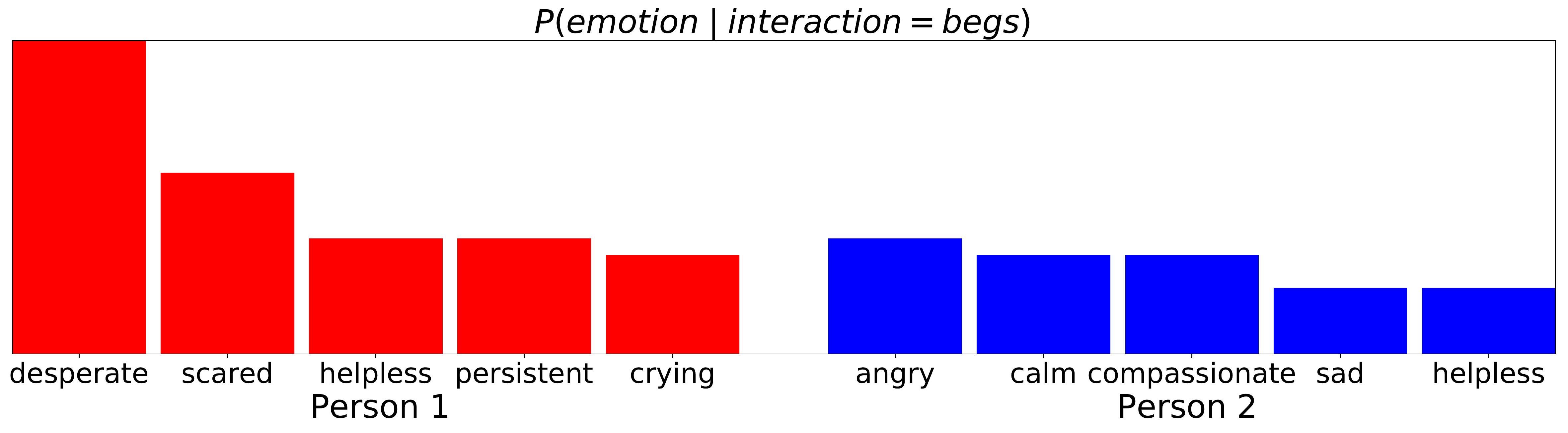} \\
    \vspace{1mm}
    \includegraphics[width=0.67\linewidth]{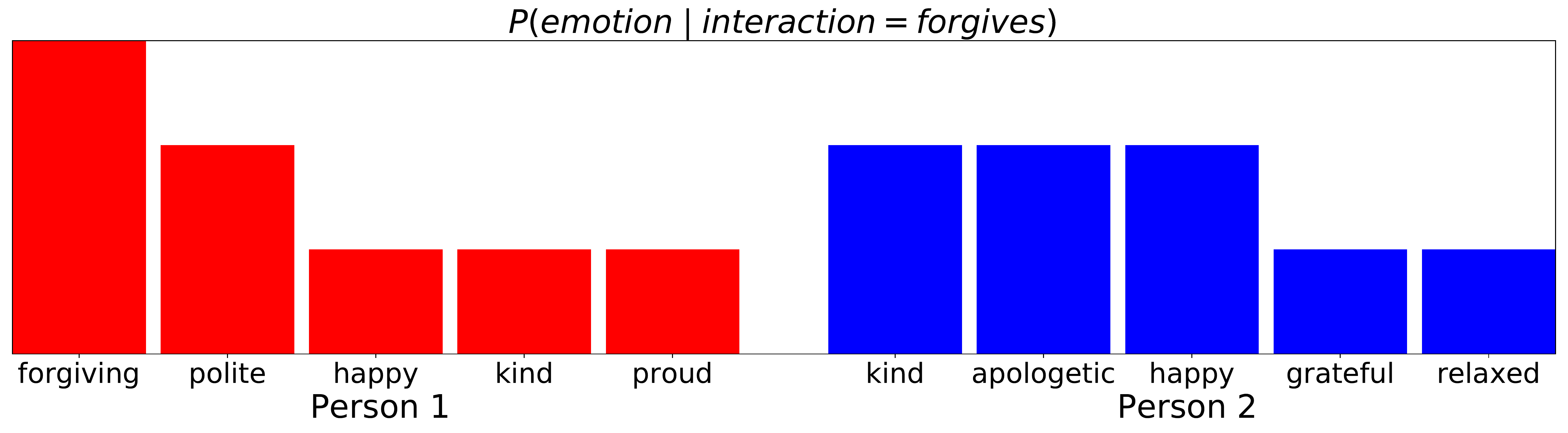} \\
\caption{\small Emotions of characters on either side of an interaction. In each case, Person 1 directs the interaction toward Person 2 (e.g. in the top example, Person 1 \textit{attacks}, Person 2 \textit{is being attacked}).}
\label{fig:personpersoninteraction}
\end{figure*}

\begin{figure*}[h]
\centering
    \includegraphics[width=0.67\linewidth]{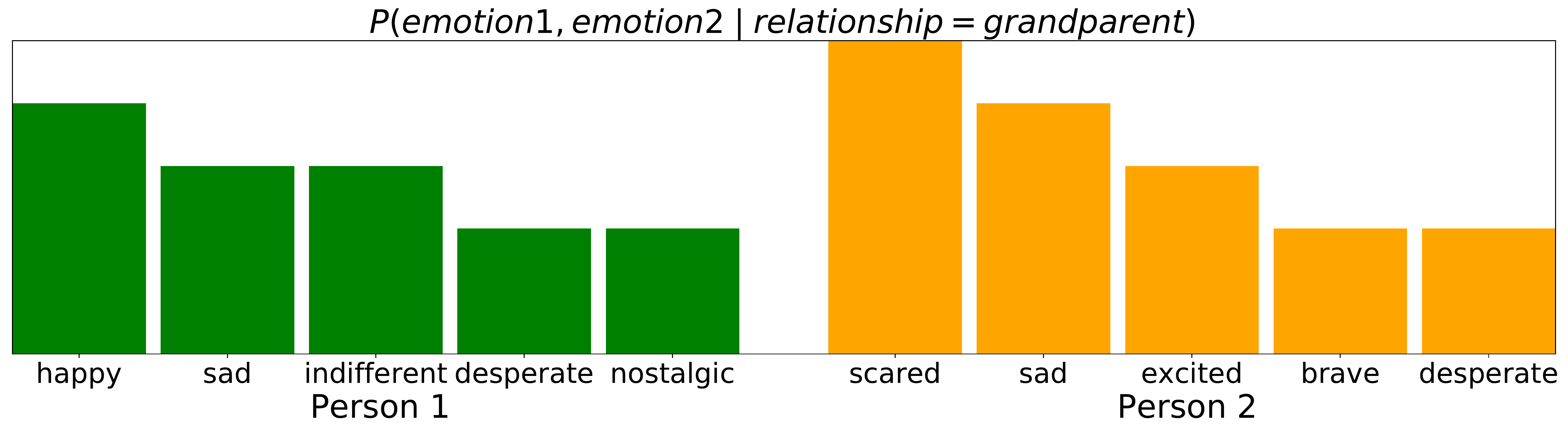} \\
    \vspace{1mm}
    \includegraphics[width=0.67\linewidth]{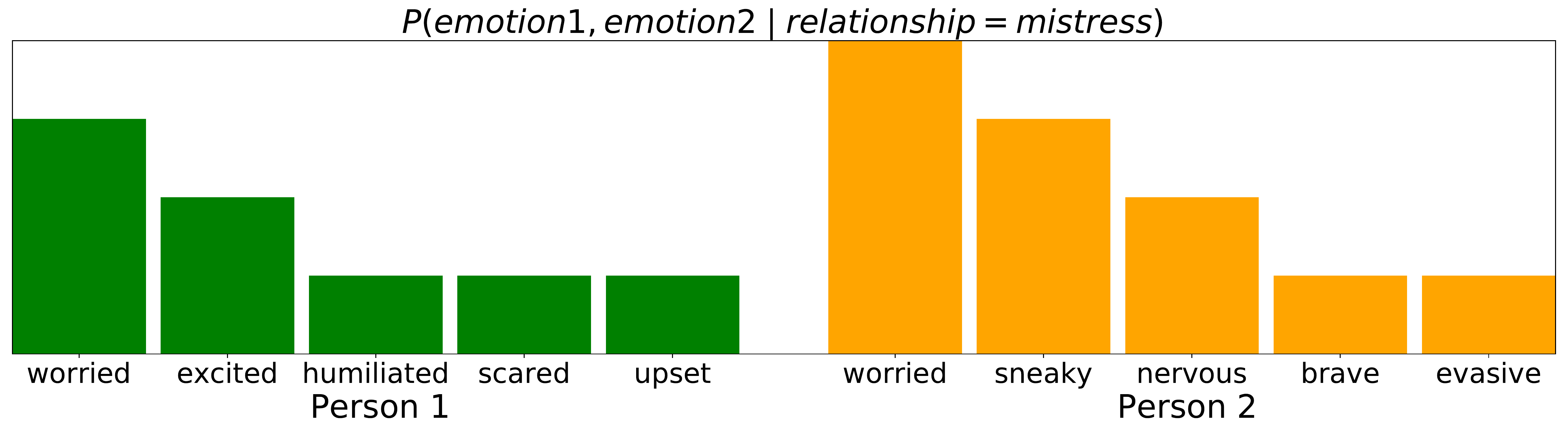} \\
    \vspace{1mm}
    \includegraphics[width=0.67\linewidth]{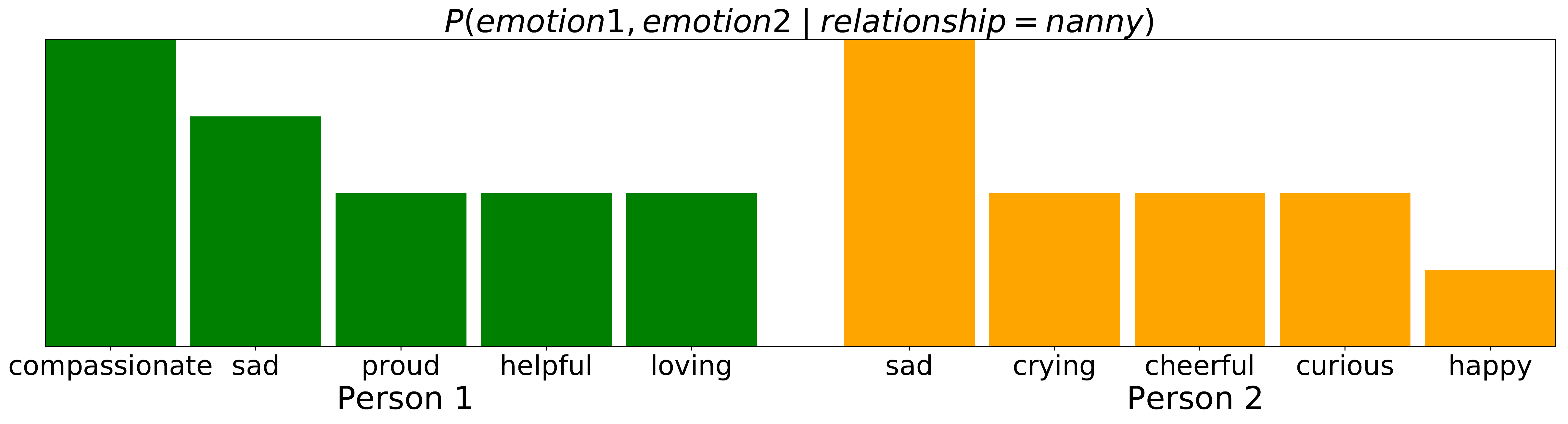} \\
\caption{\small Emotions of characters on either side of a relationship. Each relationship is directed; in these examples, Person 1 is the grandparent, mistress, and nanny, while Person 2 is the grandchild, lover, and child, respectively.}
\label{fig:personpersonrelationship}
\end{figure*}

\paragraph{Character-Character Emotion Distributions.}
In many cases, an interaction is associated with different emotions for the ``doer'' and the ``receiver.''
We show the distribution of emotions felt by characters on the giving and receiving ends of interactions (Fig.~\ref{fig:personpersoninteraction}) and relationships (Fig.~\ref{fig:personpersonrelationship}).

As shown in the top panel of the Fig.~\ref{fig:personpersoninteraction}, for the interaction \textit{attacks}, the attacker (Person 1) is often violent, angry, and aggressive, while the person being attacked (Person 2) is often scared and confused.
The emotions of a person who begs (Fig.~\ref{fig:personpersoninteraction}, middle panel) are also different (e.g. desperate, scared, helpless) from the emotions of the person on the other end of the interaction, who is often angry, calm, or compassionate.
Fig.~\ref{fig:personpersoninteraction} (bottom panel) also shows that a person who \textit{forgives} is forgiving and happy, while the person who is \textit{forgiven} is often apologetic, happy, and grateful.

People also have different emotions depending on the relationship between them, as shown in Fig.~\ref{fig:personpersonrelationship}. For example, a grandparent is often happy, while the grandchild is often scared (top panel); a mistress is often worried, and sometimes humiliated, while her lover is worried and sneaky (middle panel); and a nanny is often compassionate, while the child is often sad (bottom panel).

\begin{figure*}[p]
\centering
    \includegraphics[width=\linewidth]{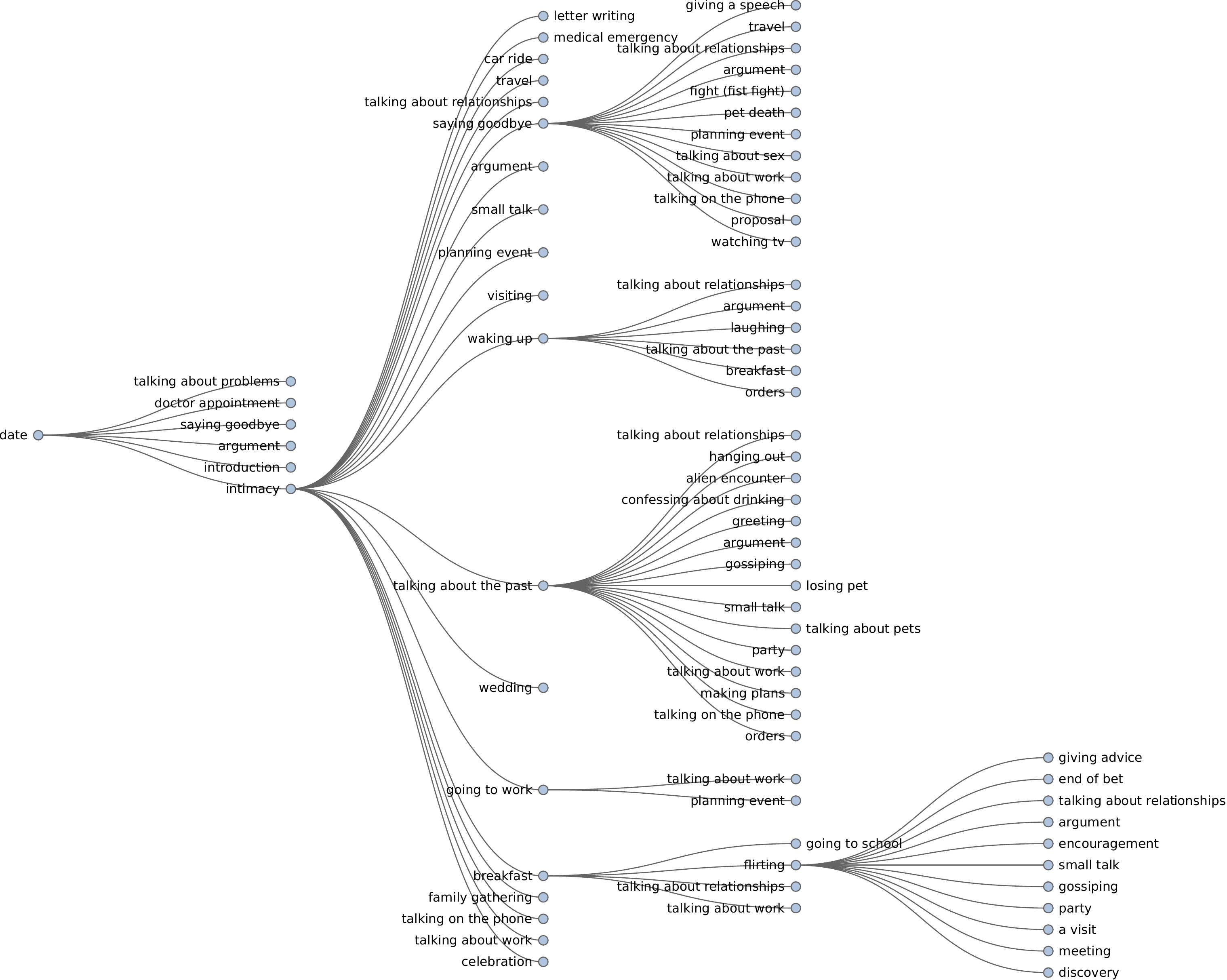}
\caption{
A tree showing possible pairwise transitions from one situation to another. For example, the situation \textit{date} can be followed by \textit{intimacy}. In turn, the situation \textit{intimacy} can be followed by, among others, \textit{talking about the past}. \textit{Talking about the past} is followed by many kinds of situations, including \textit{argument}. The sequence \textit{date}-\textit{intimacy}-\textit{talking about the past}-\textit{argument} is therefore possible, but not necessarily found in the movies we have annotated so far. This longer sequence follows from multiple pairwise transitions between situations.
}
\label{fig:situationtreedate}
\end{figure*}

\paragraph{Rooted Situation Graph.}
We can also see how situations follow from one another. Fig.~\ref{fig:situationtreedate} shows possible pairwise transitions between situations, starting from the situation ``date.''

\paragraph{Interaction Examples.}

\begin{figure*}[p]
\centering
   \includegraphics[width=0.8\linewidth]{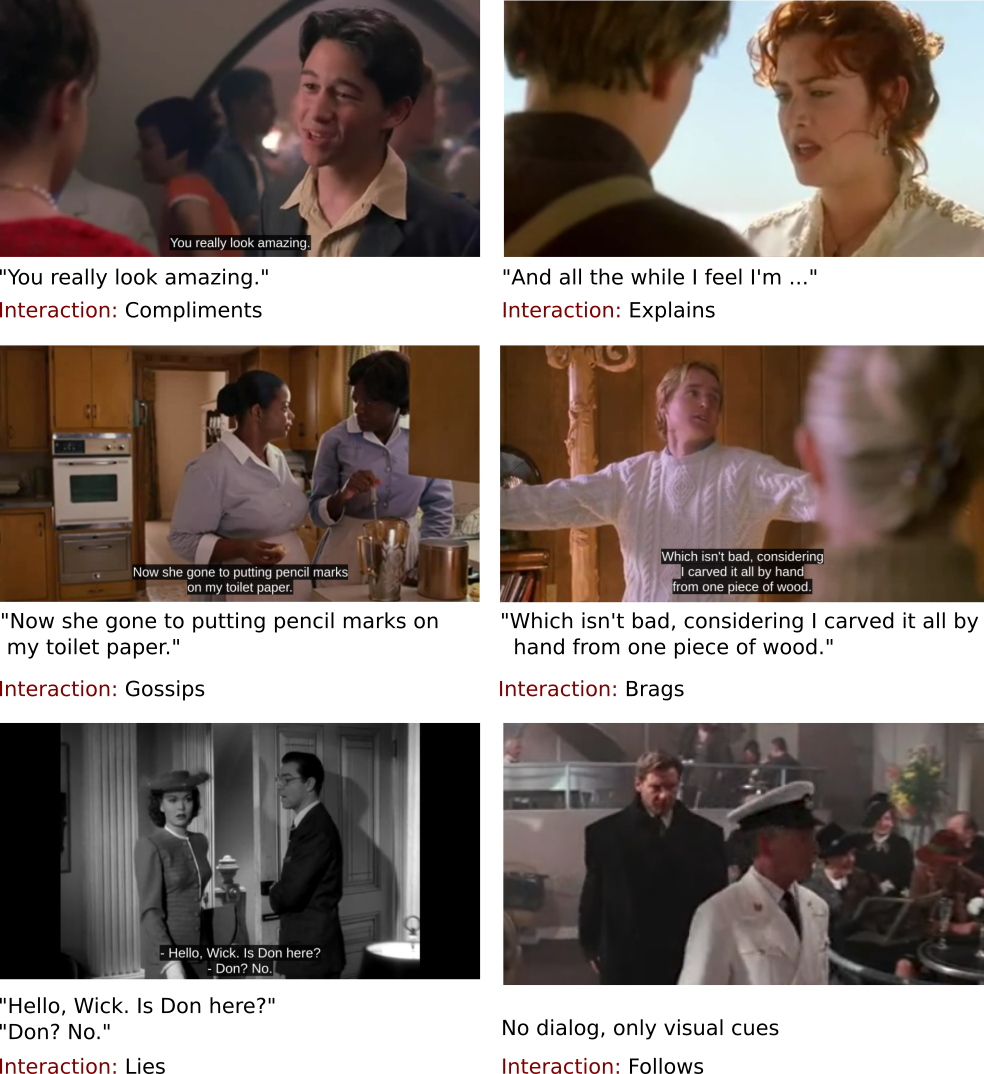}
\vspace{2mm}
\caption{\small Examples of interaction annotations. All interactions but the bottom right are inferred mainly from dialog (shown under each image); the bottom right interaction is based only on visual cues.}
\label{fig:interactionexamples}
\end{figure*}

Fig.~\ref{fig:interactionexamples} shows example annotations of interactions, to showcase the fact that dialog is important for inferring many interactions, in addition to visual cues.

\clearpage

\clearpage

\clearpage

{\small
\bibliographystyle{ieee}
\bibliography{refs}
}

\end{document}